\documentclass[sn-mathphys,iicol]{sn-jnl}% Default with double column layout

\usepackage{amssymb}
\usepackage{amsmath}
\usepackage{amsfonts}
\usepackage{amsthm}
\usepackage{mathtools}
\usepackage{bm}
\usepackage{multirow}
\usepackage{color}
\usepackage{booktabs}

\newtheorem{defi}{Definition}   
\newtheorem{pro}{Problem}    
\newcommand\figref[1]{Figure~\ref{#1}}

\newcommand\tabref[1]{Table~\ref{#1}}
\newcommand\secref[1]{Section~\ref{#1}}
\newcommand\equref[1]{Equation~(\ref{#1})}

\theoremstyle{thmstyleone}%
\theoremstyle{thmstyletwo}%

\theoremstyle{thmstylethree}%

\begin{document}

\title[Article Title]{A Correlation Information-based Spatiotemporal Network for Traffic Flow Forecasting}

%%=============================================================%%
%% Prefix	-> \pfx{Dr}
%% GivenName	-> \fnm{Joergen W.}
%% Particle	-> \spfx{van der} -> surname prefix
%% FamilyName	-> \sur{Ploeg}
%% Suffix	-> \sfx{IV}
%% NatureName	-> \tanm{Poet Laureate} -> Title after name
%% Degrees	-> \dgr{MSc, PhD}
%% \author*[1,2]{\pfx{Dr} \fnm{Joergen W.} \spfx{van der} \sur{Ploeg} \sfx{IV} \tanm{Poet Laureate} 
%%                 \dgr{MSc, PhD}}\email{iauthor@gmail.com}
%%=============================================================%%

\author[1,2]{\fnm{Weiguo} \sur{Zhu}}\email{zhuweiguo@bjtu.edu.cn}

\author*[1,2]{\fnm{Yongqi} \sur{Sun}}\email{yqsun@bjtu.edu.cn}

\author[1,2]{\fnm{Xintong} \sur{Yi}}\email{xintongyi@bjtu.edu.cn}
\author[1]{\fnm{Yan} \sur{Wang}}\email{yan.wang@bjtu.edu.cn}

\affil[1]{\orgdiv{School of Computer and Information Technology}, \orgname{Beijing Jiaotong University}, \orgaddress{\state{Beijing}, \postcode{100044}, \country{China}}}

\affil[2]{\orgname{Beijing Key Lab of Traffic Data Analysis and Mining}, \orgaddress{\state{Beijing}, \postcode{100044}, \country{China}}}
%
%\affil[3]{\orgdiv{Department}, \orgname{Organization}, \orgaddress{\street{Street}, \city{City}, \postcode{610101}, \state{State}, \country{Country}}}

%%==================================%%
%% sample for unstructured abstract %%
%%==================================%%

\abstract{The technology of traffic flow forecasting plays an important role in intelligent transportation systems. Based on graph neural networks and attention mechanisms, most previous works utilize the transformer architecture to discover spatiotemporal dependencies and dynamic relationships. However, they have not considered correlation information among spatiotemporal sequences thoroughly. In this paper, based on the maximal information coefficient, we present two elaborate spatiotemporal representations, spatial correlation information (SCorr) and temporal correlation information (TCorr). Using SCorr, we propose a correlation information-based spatiotemporal network (CorrSTN) that includes a dynamic graph neural network component for integrating correlation information into spatial structure effectively and a multi-head attention component for modeling dynamic temporal dependencies accurately. Utilizing TCorr, we explore the correlation pattern among different periodic data to identify the most relevant data, and then design an efficient data selection scheme to further enhance model performance. The experimental results on the highway traffic flow (PEMS07 and PEMS08) and metro crowd flow (HZME inflow and outflow) datasets demonstrate that CorrSTN outperforms the state-of-the-art methods in terms of predictive performance. In particular, on the HZME (outflow) dataset, our model makes significant improvements compared with the ASTGNN model by 12.7\%, 14.4\% and 27.4\% in the metrics of MAE, RMSE and MAPE, respectively.}


%%\pacs[JEL Classification]{D8, H51}

%%\pacs[MSC Classification]{35A01, 65L10, 65L12, 65L20, 65L70}

\maketitle

\section{Introduction}
Currently, many smart cities are making dramatic efforts to improve the performance of intelligent transportation systems (ITSs). As one of the most fundamental and crucial techniques in smart city construction, traffic flow forecasting has become a hot research topic.

% In this, predicting the traffic flow with massive and complex structural spatiotemporal data has become a top priority task.

%Since ITSs generate massive and complex structural spatiotemporal data, how to precisely predict the traffic flow with these data has become a top priority.

Traffic flow forecasting utilizes historical traffic data to predict future flow timestamps. Early works focusing on time series prediction have produced excellent results. Traditional methods, such as SVR~\cite{Drucker:1997}, SVM~\cite{Jeong:ww,Sun:2015wj} and KNN~\cite{van2012short,Luo:2019hm}, have been extensively applied to traffic forecasting. However, these methods need to identify the data characteristics and ignore the spatial features.

With the rapid development of deep learning, deep neural networks have been used to extract spatiotemporal features for traffic forecasting. The convolutional neural network (CNN) is introduced into traffic foresting tasks as an effective method to extract spatial features~\cite{Cui:2018deep,Yao:2018wy,Zhang:2018uj,Wang.20206tj}. However, these methods are coarse-graining processes utilizing CNN to capture neighbor block attributes and extract nonlinear spatial dependencies by a grid representation. 

Since grid representation cannot adequately represent the flow between sensors, graph representation is proposed to encode the elaborate relationships among sensors~\cite{yu2018spatio,li2018diffusion}. In recent years, the graph neural network (GNN), as an efficient and effective method, has gradually become the essential traffic prediction module for graph representation. GNN-based methods concentrate on the relationships among adjacent sensors with the propagate-aggregate mode~\cite{yu2018spatio,li2018diffusion,Guo.2019,Song.2020,Li_Zhu_2021,Guo.2021}. Furthermore, attention-based methods, such as~\cite{Guo.2019,Song.2020,Li_Zhu_2021,Guo.2021,Liu.2021,Wang.2021,Zhou.2021}, form the feature extraction processing as a query within queries (Q) and keys (K) and calculate attention weights as interrelationships for values (V).

Nevertheless, three key issues need to be given more attention. First, GNN-based methods cannot construct correct features with a sparse similarity matrix by the propagate-aggregate mode. Since the sparse similarity matrix is generated based on the spatial sensors and road, the neighboring sensors typically do not have a similar pattern. Researchers have made efforts to take mixed-hop propagation to explore deep neighborhoods~\cite{Wu:2020}, take the DTW algorithm to construct a graph matrix~\cite{Li_Zhu_2021,Fang:2021}, and take dynamic graph convolution to capture dynamic relationships~\cite{Zheng.2020,Han:2021}. Even though these methods improve traffic forecasting accuracy, their performance could be further improved if they take into account more elaborate and density correlation information.

\begin{figure}[!h]
\centering
\includegraphics[width=\columnwidth]{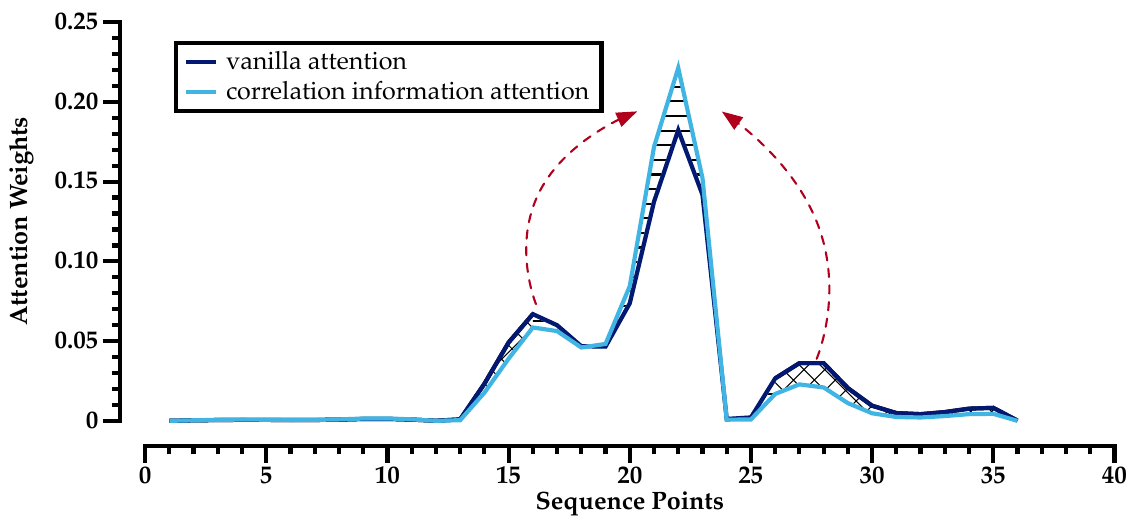}
\caption{The attention weight distribution of the vanilla attention and correlation information attention mechanisms.}
\label{fig:CIATT-diff}
\end{figure}

Second, the attention mechanism makes unwanted distractions during the attention weight calculation, and it cannot rectify the false activations induced by distractions of keys (K). As shown in \figref{fig:CIATT-diff}, the most relevant sequence part distracts attention weights to the other parts with vanilla attention.

Third, for periodic data, it is challenging to select appropriate data as input for neural network-based methods. The traditional methods iterate over all possible schemes and choose the best scheme. However, these methods are not practicable for neural network-based methods due to the expensive computation cost.

In this paper, to solve the above issues, we first introduce a spatial correlation representation (\textbf{SCorr}) to elaborately reflect the relevant features among temporal sequences. Based on SCorr, a dynamic correlation information GNN (\textbf{CIGNN}) is designed to integrate the spatial graph network with SCorr and capture more comparable patterns from each sensor. In addition, to handle the unwanted distractions in the attention mechanism, we propose a correlation information multi-head attention mechanism (\textbf{CIATT}), which can stabilize contextual features by aggregating the top-$k$ comparable patterns and concentrate attention weights on the most relevant sequence as matching templates (see \figref{fig:CIATT-diff}). Then, for spatiotemporal features, we propose an effective correlation information-based spatiotemporal network for traffic flow forecasting (\textbf{CorrSTN}), which employs CIGNN to discover the dynamic spatial dependencies and CIATT to match the relevant temporal traffic patterns. Finally, to avoid the traditional exhaustive search for the data selection scheme, we build a temporal correlation information representation (\textbf{TCorr}) to mine the relevant sequence among different periodic data. By TCorr, we design an appropriate data selection scheme to further enhance model performance.

For traffic flow forecasting, we summarize three key contributions of our work as follows:
\begin{itemize}
    \item We propose an elaborate and dense spatial correlation information representation to fully express the similarity among each sensor.
    \item In our model, the proposed CIGNN can construct correct features by associating the density similarity representation and predefined graph structure with the propagate-aggregate mode. Meanwhile, the CIATT is developed to stabilize contextual features and concentrate attention weights on the most relevant sequence.
    \item A temporal correlation information representation (TCorr) is proposed to mine the relevant sequence and seek the best data scheme. To the best of our knowledge, the efficient data selection scheme is the first used for neural network-based methods in traffic forecasting tasks.
\end{itemize}

The experimental results on four real-world datasets show that the predictive performance of CorrSTN outperforms the state-of-the-art model.
{\color{black}The remainder of this paper is organized as follows. Section 2 gives a brief overview of the related work. Section 3 formulates the traffic flow forecasting problem. Section 4 presents a detailed description of correlation information representations and the proposed model. Section 5 presents and analyzes the evaluation results. Section 6 provides detailed experiments about model settings. Section 7 gives the case studies. Section 8 summarizes our work.}

\section{Related Work}
\subsection{Traditional Traffic Forecasting Methods}
The early works for traffic forecasting are based on machine learning methods. 
Drucker et al.~\cite{Drucker:1997} propose a linear support vector machine (SVR) to predict traffic flow data. 
Lu et al.~\cite{Lu:2003br} propose an advanced time series model based on vector autoregression (VAR) to capture the pairwise relationships among spatial sequences on traffic data. 
Hochreiter and Schmidhuber~\cite{Hochreiter:1997} design the long short-term memory (LSTM) network, a special RNN, to predict time series data and solve the vanishing gradient problem.
With only temporal sequences considered and spatial relationships ignored, their prediction accuracy cannot satisfy the practical requirements.

\subsection{Deep Neural Network Traffic Forecasting Methods}
The deep neural network has dramatically improved traffic prediction accuracy by combining CNN and RNN variants. 
Li et al.~\cite{li2018diffusion} propose a diffusion convolutional recurrent neural network (DCRNN) to employ a diffusion graph convolutional network and a gated recurrent unit (GRU) in seq2seq to predict traffic data.
Yu et al.~\cite{yu2018spatio} design a spatial-temporal graph convolutional network (STGCN) architecture for spatiotemporal datasets in the traffic forecasting task.
Wu et al.~\cite{Wu:2019} propose Graph WaveNet (GWN), which combines a graph convolution network with a temporal convolution network to capture spatial-temporal dependencies. 

Since the attention mechanism can effectively model the dependencies among sequences, many works utilize it in traffic flow forecasting tasks.
Guo et al.~\cite{Guo.2019} present an ASTGCN model that uses spatial and temporal attention mechanisms to improve prediction accuracy.
Song et al.~\cite{Song.2020} propose a spatial-temporal synchronous graph convolution network (STSGCN) to extract temporal adjacency features by considering the local spatiotemporal relation. 
Recently, Guo et al.~\cite{Guo.2021} propose a GNN-based model ASTGNN formed as an encoder-decoder architecture~\cite{Vaswani.2017} with residual connection~\cite{He:2016} and layer normalization~\cite{ba2016layer} to learn the dynamics and heterogeneity of spatial-temporal graph data for traffic forecasting, which is an extension of their previous ASTGCN model.
Its prediction accuracy is significantly improved by effectively capturing the local data trend and dynamically aggregating the spatial features.

Although these methods have yielded outstanding results, they do not take the crucial correlation information among spatiotemporal sequences into account.

\subsection{Correlation Information Traffic Forecasting Methods}
The GWN adopts an adaptive adjacency matrix as a supplement to the structural matrix by calculating the vector correlations during the training stage~\cite{Wu:2019}. However, it cannot accurately reflect the correlation information by short-term sequences. 
Zheng et al.~\cite{Zheng.2020} design a graph multi-attention network (GMAN) integrating spatial and temporal attention to capture sensor correlations. 
Bai et al.~\cite{Bai:2020} propose adaptive graph generation to dynamically generate the graph during training in AGCRN.
Wu et al.~\cite{Wu:2020} make use of the graph learning layer in the MTGNN to construct an adaptive graph by multivariate node features.
Li and Zhu~\cite{Li_Zhu_2021} introduce a spatial-temporal fusion graph matrix combining the correlation information with the structural network (STFGNN). However, since the matrix is very sparse, the correlation among the sensors is not elaborated for traffic forecasting.
Fang et al.~\cite{Fang:2021} combine the spatial and semantical neighbors to consider spatial correlations in STGODE.
Han et al.~\cite{Han:2021} propose a dynamic graph constructor and graph convolution in the DMSTGCN to learn the dynamic spatial dependencies as an extensive predefined adjacency matrix.

However, their methods cannot extract correct features in the GNN-based component and avoid unwanted distractions in the attention-based component. Based on the correlation information, we aim to obtain more correct features and more focused attention weights in this paper.

\begin{figure*}[t]
\centering
\includegraphics[width=\textwidth]{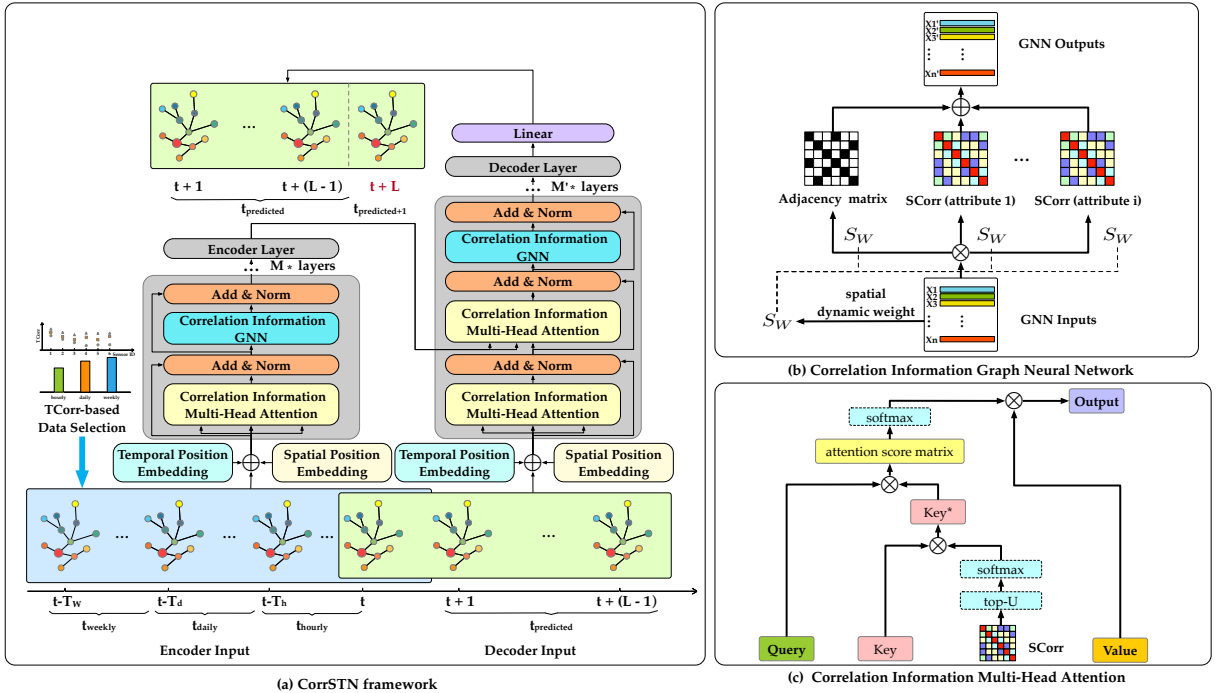}
\caption{An illustration of the proposed framework. The framework is a transformer network based on the encoder-decoder architecture. Each encoder (decoder) layer contains a (two) correlation information multi-head attention component(s) and a correlation information graph neural network component.}
\label{fig:model}
\end{figure*}

%%%%%%%%%%%%%%% Preliminaries %%%%%%%%%%%%%%%
\section{Preliminaries}\label{sec:preliminary}
\begin{defi}
Traffic Network. We define a traffic network as a directed or an undirected graph $\mathbf{G} = (\mathbf{V},\mathbf{E})$, where $\mathbf{V}$ is a set of $\|\mathbf{V}\| = N$ nodes, each node represents a traffic sensor, and $\mathbf{E}$ is a set of edges.
\end{defi}

\begin{defi}
Traffic spatiotemporal sequence. 
We define a traffic spatiotemporal sequence as 
$\mathbf{X} = (\mathbf{X}^{1}, \mathbf{X}^{2}, \ldots ,\mathbf{X}^{T}) \in \mathbb{R}^{T \times N \times C}$
,where $\mathbf{X}^{t} = (\mathbf{x}_{1}^{t}, \mathbf{x}_{2}^{t}, \ldots ,\mathbf{x}_{N}^{t}) \in \mathbb{R}^{N \times C}$ denotes the vector of the $N$ sensors with $C$ attributes at timestamp $t$.
\end{defi}

\begin{defi}
 Periodic Data.
We define the hourly, daily and weekly data intervals as $T_{h}$, $T_{d}$ and $T_{w}$, respectively. Given time window $\tau$, the historical periodic data can be defined as
\begin{equation}
\label{equ:input}
    \begin{aligned}
        \mathbf{X} = (&\mathbf{X}^{t-T_{w}+1}, \mathbf{X}^{t-T_{w}+2}, \ldots, \mathbf{X}^{t-T_{w}+\tau},\\
                      &\mathbf{X}^{t-T_{d}+1}, \mathbf{X}^{t-T_{d}+2}, \ldots, \mathbf{X}^{t-T_{d}+\tau},\\
                      &\mathbf{X}^{t-T_{h}+1}, \mathbf{X}^{t-T_{h}+2}, \ldots, \mathbf{X}^{t-T_{h}+\tau}),
    \end{aligned}
\end{equation}
where the time interval of each timestamp is 5 minutes on the datasets, and $\tau = 12$ in this paper. 
\end{defi}

\begin{pro}
Given the historical periodic data $\mathbf{X} \in \mathbb{R}^{T_{hdw} \times N \times C}$ defined as~\equref{equ:input}, where $T_{hdw} \in [\tau,2\tau,3\tau]$ will change according to our data selection scheme for different datasets. Then our focus is to predict traffic flow for all sensors over the next $\mathbf{L}$ timestamps,
\begin{equation}
    f(\mathbf{X}) \rightarrow (\hat{\mathbf{X}}^{t+1}, \hat{\mathbf{X}}^{t+2}, \ldots, \hat{\mathbf{X}}^{t+\mathbf{L}}) \in \mathbb{R}^{L \times N \times 1},
\end{equation}
where $f(\cdot)$ is the mapping function aimed at learning and $\mathbf{L} = 12$ in our model.
\end{pro}

%%%%%%%%%%%%%%% Method %%%%%%%%%%%%%%%
\section{Methodology}
In this section, we will introduce our spatiotemporal correlation information representations (SCorr and TCorr) and the correlation information-based components (CIGNN and CIATT). The overall framework of CorrSTN is based on an encoder-decoder architecture, as shown in~\figref{fig:model}. The encoder (decoder) network consists of temporal position embedding, spatial position embedding and encoder (decoder) layer components. The CIGNN and CIATT components are connected by the residual connection and layer normalization in each encoder and decoder layer. The second CIATT component of each decoder layer is designed to receive the encoder output as historical memory.

\subsection{Spatiotemporal Correlation Information Representation}
In this subsection, we propose two elaborate spatiotemporal representations based on the maximal information coefficient~\cite{Reshef.2011}, spatial correlation information (SCorr) and temporal correlation information (TCorr). To measure the degree of correlation information among sensor sequences, the maximal information coefficient (MIC) method is employed in the two spatiotemporal representations. Compared with DTW and cosine similarity methods, MIC can not only capture diverse associations but also execute fast calculations.

To explain the calculation process of MIC in detail, we first define two sequences, $var_1 \in \mathbb{R}^{M}$ and $var_2 \in \mathbb{R}^{M}$. By partitioning the $x$-axis into $\mathbf{A}$ parts and the $y$-axis into $\mathbf{B}$ parts, the degree of correlation information between $var_1$ and $var_2$ can be calculated as follows,
\begin{equation}
    \mathbf{MIC}(var_1,var_2) =
    \frac{ \max_{A*B<M^\eta}
          \left\{\textbf{I}_{a,b}(var_1, var_2)\right\}
        }
        {log_2\min\{A,B\}},
\end{equation}
where $\eta$ is a parameter to control the number of partitions and $\mathbf{I}_{a,b}$ denotes the mutual information. $\mathbf{I}_{a,b}$ is calculated as follows,
$$\mathbf{I}_{a,b} = \sum_{a<A, b<B} q(a,b) log_2 \left(\frac{q(a,b)}{q(a)q(b)}\right),$$
where $q(a,b)$ is the joint probability density and $q(a)$ and $q(b)$ are the edge probability densities when choosing the $(a, b)$ grids.

\subsubsection{Spatial Correlation Information}
In traffic forecasting tasks, the input data sequence shows interrelated spatial characteristics. In this subsection, we propose SCorr to represent the dependence among spatiotemporal sensor sequences.

Let $X_{i}^{c}$ denote the sensor $i$ vector with attribute $c$ of $T$ timestamps. Then, SCorr is defined as follows,
\begin{equation}
\label{equ:SCorr}
\mathbf{SCorr}(X)_{i,j}^{c} = \mathbf{MIC}(X_i^c,X_j^c),
\end{equation}
where $\mathbf{SCorr}(X) \in \mathbb{R}^{N \times N \times C}$ is the degree of correlation information, and $\mathbf{SCorr}(X)_{i,j}^{c} \in [0, 1]$ denotes the degree of correlation information between sensor $i$ and sensor $j$ in attribute $c$. 

\begin{figure}[h]
\centering
\includegraphics[width=\columnwidth]{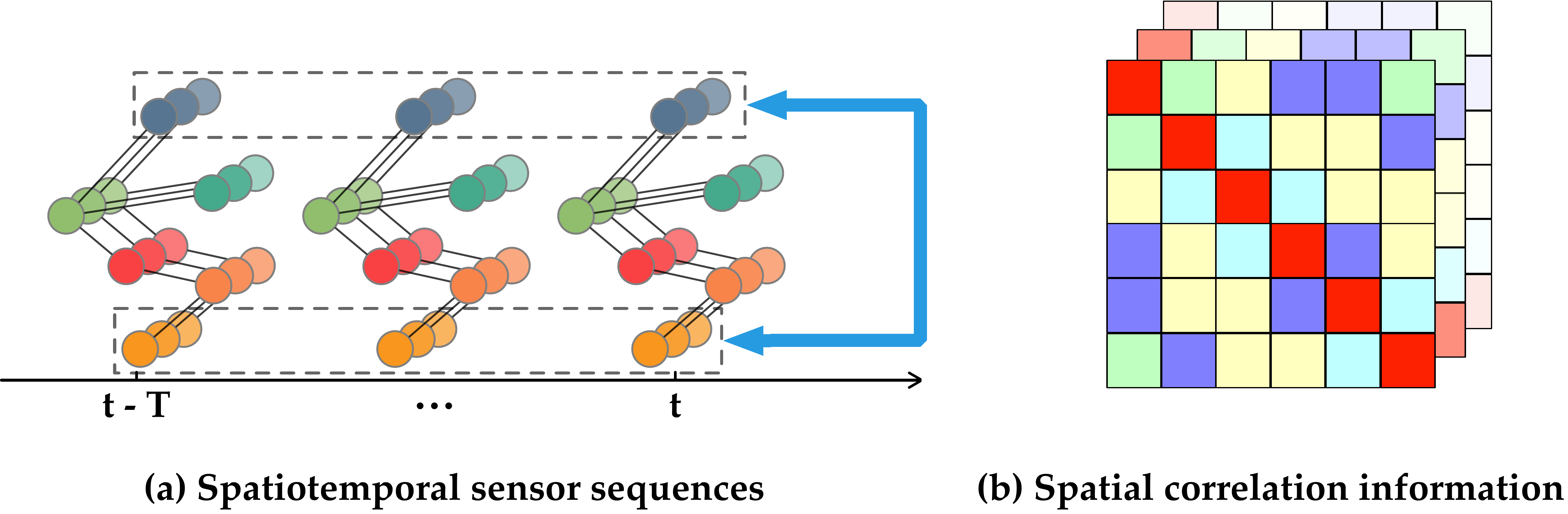}
\caption{An illustration of the SCorr calculation.}
\label{fig:SCorr}
\end{figure}

Concretely, as shown in~\figref{fig:SCorr}(a), the data sequences with $T$ timestamps are displayed. We select the upper and lower sensor sequences to calculate the degree of correlation information as one example. According to~\equref{equ:SCorr}, we calculate the degree for each sensor pair, and then the SCorr matrices are shown in~\figref{fig:SCorr}(b).

Generally, each sensor has varying correlative relationships with other sensors in the traffic road or station network. The sparse adjacency matrix cannot represent it just by 1 or 0. In contrast, the SCorr matrices are elaborate and density spatiotemporal representations with values ranging from 0 to 1. High values signify strong correlative relationships and similar patterns between sensors, while low values express significant differences and low reference characteristics.

\subsubsection{Temporal Correlation Information}
\label{sec:TCorr-define}
It has been found that traffic flow data have different temporal associations among different temporal sequences. In this subsection, we propose an effective representation, called TCorr, to explore and capture similar patterns among different periodic data.

\begin{figure}[h]
\centering
\includegraphics[width=\columnwidth]{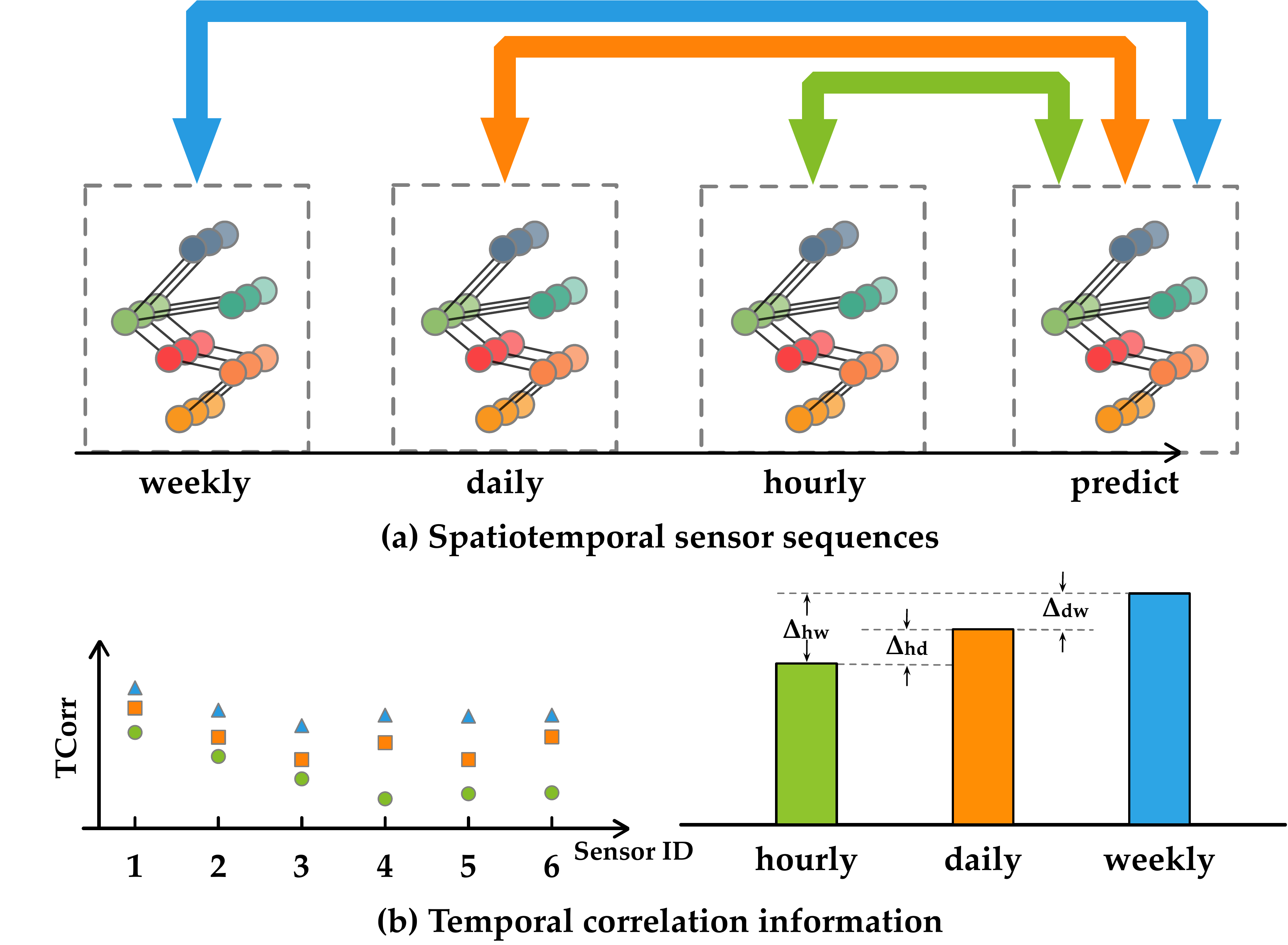}
\caption{An illustration of the TCorr calculation.}
\label{fig:TCorr}
\end{figure}

We define $X_{hourly} \in \mathbb{R}^{\tau \times N \times C}$ (resp. $X_{daily}$, $X_{weekly}$) as the vector of the last hour (resp. day, week) before the predicted data $\widetilde{X} \in \mathbb{R}^{\tau \times N \times C}$. Then, we define TCorr as follows,
\begin{equation}
    \mathbf{TCorr}(X)_{i}^{c} = \frac{1}{T}\sum\nolimits_{t=1}^{T}\mathbf{MIC}(X_{i,t}^{c,t+\tau}, \widetilde{X}_{i, t}^{c,t+\tau}),
\end{equation}
where $\mathbf{TCorr}(X) \in \mathbb{R}^{N \times C}$ is the average degree of all sensors temporal correlation information, $\mathbf{TCorr}(X)_{i}^{c} \in [0, 1]$ denotes the degree of temporal correlation information of sensor $i$ in attribute $c$, and $X_{i,t}^{c,t+\tau}$ denotes the vector of sensor $i$ in attribute $c$ between timestamp $t$ and timestamp $t+\tau$ of $X_{hourly}$, $X_{daily}$ or $X_{weekly}$. We set a lower weight for timestamps shifted with a greater number of steps. Here, we set different weights for different periodic types of data as follows,
\begin{equation}
\label{equ:TCorr}
    \begin{aligned}
            \mathbf{TCorr}_{h}  = & \alpha \mathbf{TCorr}(X_{hourly}) \\
            \mathbf{TCorr}_{d}  = & \beta  \mathbf{TCorr}(X_{daily})  \\
            \mathbf{TCorr}_{w}  = & \gamma \mathbf{TCorr}(X_{weekly}),\\
    \end{aligned}
\end{equation}
where the weights $\alpha$, $\beta$ and $\gamma$ are set at 0.95, 0.95 and 0.85 according the shifted steps of hour, day and week in this paper, respectively. We illustrate the TCorr calculation between the last hour (day, week) data and the predicted data, as shown in~\figref{fig:TCorr}(a). 

The previous neural network-based methods take the data selection scheme as hyperparameters and exhaustively search for the appropriate scheme to improve the model performance. In the following, with the help of TCorr, we will design an efficient data selection scheme.

%In this subsection, for the selection data challenge, we propose an efficient method to search for the appropriate data selection scheme by using TCorr.

%we can assess the contribution of each periodic data properly.

\textbf{Data selection scheme.} First, we evaluate the contributions of different periodic data for the predicted data. With the help of TCorr, we define $\Delta_{hd}$, $\Delta_{hw}$ and $\Delta_{dw}$ to represent the contributions as follows, 
\begin{equation}
    \begin{aligned}
            \Delta_{hd}^c  = & \frac{1}{N}\sum\nolimits_{i=1}^N \mathbf{TCorr}_{d,i}^c - \frac{1}{N}\sum\nolimits_{i=1}^N \mathbf{TCorr}_{h,i}^c  \\
            \Delta_{hw}^c  = & \frac{1}{N}\sum\nolimits_{i=1}^N \mathbf{TCorr}_{w,i}^c - \frac{1}{N}\sum\nolimits_{i=1}^N \mathbf{TCorr}_{h,i}^c \\
            \Delta_{dw}^c  = & \frac{1}{N}\sum\nolimits_{i=1}^N \mathbf{TCorr}_{w,i}^c - \frac{1}{N}\sum\nolimits_{i=1}^N \mathbf{TCorr}_{d,i}^c, \\
    \end{aligned}
\end{equation}
where $\Delta_{hd}^c$ (resp. $\Delta_{hw}^c$, $\Delta_{dw}^c$) denotes the gap of contribution between hourly and daily (resp. hourly and weekly, daily and weekly) data with attribute $c$. To clearly, we illustrate the $\mathbf{TCorr}$ of all sensors, as the scatter chart shows in~\figref{fig:TCorr}(b), while the $\Delta_{hd}^c$, $\Delta_{hw}^c$ and $\Delta_{dw}^c$ are notated on the average values, as the histogram shows in~\figref{fig:TCorr}(b).

%decomposition
%{\color{blue}Additionally, time-series data can be decomposed into trend data (short-term data) and season data (long-term data)~\cite{Williams:2003br}. It is also widely applicable for traffic flow data.

Then, according to the data contribution, we are able to make the appropriate data selection scheme by the following three rules: i) to capture the short-term tendency, we set the hourly data as the basis of the input data; ii) the daily (weekly) data are combined into input data if $\Delta_{hd}^c>0$ ($\Delta_{hw}^c>0$) to capture the long-term season; iii) according to Occam's Razor principle, when $\Delta_{hd}^c>0$ and $\Delta_{hw}^c>0$, we select the periodic data to combine with input data as follows,
\begin{equation}
    \label{equ:eight}
    \text {data} = 
    \begin{cases}
    \text {daily and weekly data}, & \text {for} \  \Delta_{dw}^c>0, \\
    \text {daily or weekly data},  & \text {for} \  \Delta_{dw}^c=0, \\
    \text {daily data},            & \text {for} \  \Delta_{dw}^c<0,
    \end{cases}
\end{equation}
where the equation $\Delta_{dw}^c>0$ denotes that data have weekly periodic properties. 
As shown in~\figref{fig:TCorr}(b), we can see that $\Delta_{hd}^c > 0$, $\Delta_{hw}^c>0$ and $\Delta_{dw}^c>0$. Thus, the appropriate scheme selects the hourly, daily and weekly data as input data for the example.

Based on our three rules and~\equref{equ:eight}, we can search for an appropriate data selection scheme for datasets of various types. Thus, we can make use of the relevant sequence among different periodic data to further enhance model performance. Additionally, we will give more TCorr representations of different datasets and discuss different schemes in~\secref{sec:TCorr} in depth.

\subsection{Correlation Information Graph Neural Network}
In this subsection, we propose a correlation information graph neural network (CIGNN) incorporating the spatial correlation information into the structure network, as shown in \figref{fig:model}(b). The vanilla graph neural network is defined as follows,
\begin{equation}\label{equ:gcn}
\mathbf{Z}^{(l)} = \sigma\left(\mathbf{A} \mathbf{Z}^{(l-1)} \mathbf{W}^{(l)}\right),
\end{equation}
where $\mathbf{Z}^{(l)}$ and $\mathbf{Z}^{(l-1)} \in \mathbb{R}^{N \times d_{model}}$, $\mathbf{W}^{(l)} \in \mathbb{R}^{d_{model} \times d_{model}} $, and $\sigma$ are the sensor feature representation output and the input of $d_{model}$ dimensions, linear projection weight matrix, and nonlinear activation function, respectively.
$l$ and $l-1$ denote the layer number.
$\mathbf{A} \in \mathbb{R}^{N \times N}$ denotes the normalized structural adjacency matrix by the Laplacian regularization $$\mathbf{A} = \widetilde{\mathbf{D}}^{-\frac{1}{2}} \widetilde{\mathbf{A}} \widetilde{\mathbf{D}}^{-\frac{1}{2}},$$
where $\widetilde{\mathbf{A}} \in \mathbb{R}^{N \times N} $ is the graph adjacency matrix and $\widetilde{\mathbf{D}}$ is the diagonal matrix with the $i_{th}$ element $\widetilde{\mathbf{D}}_{ii} = \sum_j \widetilde{\mathbf{A}}_{ij}$.

According to \equref{equ:gcn}, the network only considers the neighbor sensor relationships, even if the neighbor sensors do not have similar features and patterns. Aiming to widely capture similar patterns from other sensor nodes, we propose the CIGNN module as follows: 
\begin{equation}\label{equ:new-gcn}
        \mathbf{\hat{Z}}^{(l)} =\textbf{AGG}\left( \psi_{c} \sigma \left(\mathbf{SCorr}_{c} \mathbf{S}_{w} \mathbf{Z}^{(l-1)} \mathbf{W}^{(l)}\right) \right),
\end{equation}
where $\mathbf{SCorr}_{c} \in \mathbb{R}^{N \times N}$ denotes our proposed spatial correlation information in dimension $c$, and $\psi_c$ is the trainable parameter to control the aggregation level of each attribute. 

To cover the dynamic change among sensors over time, we add a spatial dynamic weight matrix $\mathbf{S}_{w}$ to adaptively adjust the degree of correlation information as in~\cite{Guo.2021}. The spatial dynamic weight matrix is calculated as follows,
\begin{equation}\label{equ:st}
\mathbf{S}_{w}=\operatorname{softmax}\left(\frac{\mathbf{Z}^{(l-1)} \mathbf{Z}^{(l-1)^{T}}}{\sqrt{d_{model}}}\right) \in \mathbb{R}^{N \times N}.
\end{equation}
Although Scorr is static based on the dataset distribution, we can make use of $\mathbf{S}_{w}$ to dynamically control the feature construction. To discuss the effort of dynamic and static Scorr, we present detailed experiments in \secref{sec:sta-dyn}.

Finally, we integrate the predefined graph structure to avoid losing local information by a trainable parameter $\Omega$ as follows,
\begin{equation}\label{equ:all-gcn}
\mathbf{\hat{Z}}^{(l)}= \textbf{AGG}(\mathbf{\hat{Z}}^{(l)}, \Omega\mathbf{Z}^{(l)}).
\end{equation}

Compared with previous works, the CIGNN module integrates correlation information with structural information and provides accurate and dense interconnections for the propagate-aggregate mode. As a result, it can improve the feature aggregation efficiency among similar sensors and produce more correct features for other components.

%%%%%%%%%%%%%%%%%%%%%%%%%%%%%%%%%%%%%%%%%%%%%%%%%%%%%%%%%%%%
\subsection{Correlation Information Multi-Head Attention}
In this subsection, we propose a correlation information multi-head attention (CIATT) component utilizing spatial correlation information to construct more stable contextual features and focus on the most relevant sequence, as shown in \figref{fig:model}(c).

Currently, many works use the attention mechanism for traffic flow forecasting. The attention mechanism, as the major component of transformer architecture, utilizes the queries (\textbf{Q}), keys (\textbf{K}), and values (\textbf{V}) as the inputs to model the dependencies of time points among sequences and construct a new embedding representation output as follows,
\begin{equation}\label{equ:attn}
    \operatorname{Attention}(\mathbf{Q}, \mathbf{K}, \mathbf{V}) = \operatorname{Softmax}(\frac{\mathrm{\textbf{Q}\textbf{K}}^{T}}{\sqrt{d_{head}}}) \mathbf{V},
\end{equation}
where $\mathbf{Q} \in \mathbb{R}^{N \times L_{Q} \times d_{head}}$, $\mathbf{K} \in \mathbb{R}^{N \times L_{K} \times d_{head}}$, $\mathbf{V} \in \mathbb{R}^{N \times L_{V} \times d_{head}}$ and $d_{head}$ is the input dimension of one head. 

%\begin{equation}\label{equ:attn-final}
%    \mathrm{CIATT} (\mathbf{Q}, \mathbf{K}, \mathbf{V})  = \mathrm{Linear}(\mathrm{Cat}(\mathrm{CIATT}_1,\dots,\mathrm{CIATT}_H)),
%\end{equation}

However, unwanted distractions will appear caused by distractions of keys (K) in \equref{equ:attn}. To address the problem, we define the CIATT as follows,
\begin{equation}\label{equ:attn-final}
\begin{aligned}
    & \mathrm{CIATT} (\mathbf{Q}, \mathbf{K}, \mathbf{V}) \\
    &= \mathrm{Linear}(\mathrm{Cat}(\mathrm{CIATT}_1,\dots,\mathrm{CIATT}_H)),
\end{aligned}
\end{equation}
where there are $H$ heads for the CIATT, and the outputs of all heads are concatenated and reweighted by the Cat and Linear operators. The $h_{th}$ head is defined as follows,
\begin{equation}
\mathrm{CIATT}_h = \mathrm{Softmax}\left(\frac{\mathbf{Q}_{(h)} \widetilde{\textbf{K}}_{(h)}^{T}} {\sqrt{d_{head}}}\right) \mathbf{V}_{(h)}.
\end{equation}

Considering that different sensors with strong correlation information have similar patterns in the traffic road network, we construct $\widetilde{\textbf{K}}_{(h)}$ by using SCorr to combine similar patterns of correlative sensors as follows,
\begin{equation}\label{equ:reconstruct}
    \widetilde{\textbf{K}} = \mathcal{F}(\mathbf{SCorr},\mathbf{K})
                    = \frac{1}{C}  \sum\nolimits_{c=1}^{C} \mathbf{\hat{SCorr}}_{c}^{T}  \mathbf{\hat{K}},
\end{equation}
where $\mathcal{F}$ is the reconstruction function. The inputs are the spatial correlation information and the original keys $\mathbf{K}$ and the outputs $\widetilde{\textbf{K}} \in \mathbb{R}^{N \times L_{\mathbf{K}} \times d_{head}}$ are more stable and reliable representations for the attention mechanism, where $L_{\mathbf{K}}$ is the length of $\textbf{K}$. According to the degree of correlation information, the top $U$ items are selected from $\mathbf{SCorr}$ and normalized via the softmax function as $$ \mathbf{\hat{SCorr}}_{i,m}^{c} = \frac{exp(\textbf{SCorr}_{i,m}^{c})}{\sum_{u=1}^{U} exp(\textbf{SCorr}_{i,u}^{c})},$$ and $\hat{\mathbf{K}} \in \mathbb{R}^{N \times U \times L_\textbf{K} \times d_{head}}$ are the items selected from $\mathbf{K}$ in the same order.
In detail, the $\widetilde{\textbf{K}}_i$ of sensor $i$ in $\widetilde{\textbf{K}}$ is calculated as follows, 
$$\widetilde{\textbf{K}}_i = \frac{1}{C} \sum_{c=1}^{C} \sum_{u=1}^{U} \hat{\textbf{SCorr}}_{i,u}^{c} \hat{k}_{i,u},$$ 
where $\hat{k}_{i,u} \in \mathbb{R}^{L_\textbf{K} \times d_{head}}$ is the element of sensor $i$ in $\mathbf{\hat{K}}$. Finally, the representation $\widetilde{\textbf{K}} \in \mathbb{R}^{N \times L_{\mathbf{\textbf{K}}} \times d_{head}}$ can replace the original keys ($\textbf{K} \in \mathbb{R}^{N \times L_{\mathbf{\textbf{K}}} \times d_{head}}$) in the attention mechanism in \equref{equ:attn}.

Our method aggregates similar patterns of different sensors with SCorr weights to construct more stable contextual features for the attention mechanism. Hence, the attention mechanism can earn more focused attention weights, as shown in \figref{fig:CIATT-diff}. With the help of CIATT, a time-series sequence can accurately find its most relevant sequence to increase the accuracy of traffic flow forecasting tasks.

\subsection{Encoder-Decoder}
The encoder-decoder is the basic architecture of the transformer network. In this paper, the input data of the encoder are the periodic data defined as~\equref{equ:input}. The encoder input data are $\mathbf{X} = (\mathbf{X}_{t_{weekly}},\mathbf{X}_{t_{daily}},\mathbf{X}_{t_{hourly}})$, which are combined with the spatial position embedding and temporal position embedding. Then, CIATT constructs more stable contextual features and CIGNN integrates correlation information with structural information to learn the similar pattern for each sensor node, which are connected by the residual connection and layer normalization in each layer. The decoder input data are $\mathbf{X} = (\mathbf{X}_{t},\mathbf{X}_{t+1},\ldots,\mathbf{X}_{t+L-1})$, which are processed as the encoder input. Furthermore, the decoder layer receives the encoder output as historical memory for prediction. Finally, the output data are $\mathbf{X} = (\hat{\mathbf{X}}_{t+1},\hat{\mathbf{X}}_{t+2},\ldots,\hat{\mathbf{X}}_{t+L})$.

%%%%%%%%%%%%%%% Experiments %%%%%%%%%%%%%%%
\section{Experiments}
In this section, we conduct experiments to evaluate the performance of CorrSTN on four real-world spatiotemporal traffic network datasets and discuss the effect of each component. 
We partition each dataset into training, validation and test sets in a ratio of 6:2:2 by timestamps, and all data are normalized into [-1,1] by the min-max method. In SCorr and TCorr, $\eta$ is set at 0.6 to control the number of partitions. We evaluate the performance of our model by the metrics of mean absolute error (MAE), root mean square error (RMSE) and mean absolute percentage error (MAPE). They are defined as,
\begin{equation}\label{equ:metrics}
\begin{aligned}
\operatorname{MAE}  \left(\hat{\mathbf{X}}^{i}, \mathbf{X}^{i}\right)  &= \frac{1}{L} \sum\nolimits_{i=1}^{L}\left\|\mathbf{X}^{i}-\hat{\mathbf{X}}^{i}\right\| \\
\operatorname{RMSE} \left(\hat{\mathbf{X}}^{i}, \mathbf{X}^{i}\right)  &= \sqrt{\frac{1}{L} \sum\nolimits_{i=1}^{L}\left(\mathbf{X}^{i}-\hat{\mathbf{X}}^{i}\right)^{2}} \\
\operatorname{MAPE} \left(\hat{\mathbf{X}}^{i}, \mathbf{X}^{i}\right)  &= \frac{1}{L} \sum\nolimits_{i=1}^{L} \frac{\left\|\mathbf{X}^{i}-\hat{\mathbf{X}}^{i}\right\|}{\mathbf{X}^{i}}.
\end{aligned}
\end{equation}

\subsection{Parameter Settings}
    Our model chooses the MAE loss function and Adam optimizer for training. We feed the historical data into the encoder and decoder networks during training. Once the decoder network generates prediction results, the optimizer adjusts the model parameters according to the training loss. We set the learning rate at 0.001, and other hyperparameters are described in \tabref{tab:setting}. To compare all the baseline methods, we adopt three kinds of periodic traffic data (hourly, daily and weekly) to predict the traffic flow in the next hour. Then, we set hourly, daily and weekly periodic traffic data for the highway traffic flow datasets, while we set hourly and daily periodic traffic data for the metro crowd flow datasets. These data schemes are searched with the help of TCorr. To demonstrate the correctness of TCorr, we illustrate TCorr and conduct experiments to evaluate the scheme on the HZME (outflow) dataset in~\secref{sec:TCorr}.
    
\begin{table}[h]
    \centering
    \caption{Hyperparameters of our CorrSTN model for four datasets.}
    \label{tab:setting}
\resizebox{\columnwidth}{!}{
\begin{tabular}{lcccccc}
\hline
                 & \begin{tabular}[c]{@{}c@{}}\#encoder\\ layers\end{tabular} & \begin{tabular}[c]{@{}c@{}}\#decoder\\ layers\end{tabular} & \begin{tabular}[c]{@{}c@{}}\#kernel\\ size\end{tabular} & \#heads & \#batchsize & \#U \\ \hline
PEMS07           & 3                                                          & 3                                                          & 3                                                       & 8       & 4       & 5    \\
PEMS07(p)        & 3                                                          & 3                                                          & 3                                                       & 8       & 2       & 5    \\
PEMS08           & 4                                                          & 4                                                          & 3                                                       & 8       & 16      & 8    \\
PEMS08(p)        & 4                                                          & 4                                                          & 3                                                       & 8       & 8       & 5    \\
HZME(in)           & 4                                                          & 4                                                          & 3                                                       & 8       & 4       & 4    \\
HZME(in)(p)        & 4                                                          & 4                                                          & 3                                                       & 8       & 16     & 5     \\
HZME(out)          & 3                                                          & 3                                                          & 5                                                       & 4       & 8        & 5   \\
HZME(out)(p)       & 4                                                          & 4                                                          & 3                                                       & 4       & 16      & 5    \\ \hline
\end{tabular}
}
\end{table}

\subsection{Datasets}
\label{sec:datasets}
The details of the four traffic datasets are given in \tabref{tab:dataset}. The datasets of the first type are collected from two districts in California~\cite{Chen.2001}, namely, PEMS07 and PEMS08, which contain 883 and 170 nodes, respectively. The datasets of the second type are collected from the Hangzhou metro system~\cite{Guo.2021}, namely, HZME, including inflow and outflow datasets. The HZME datasets contain 80 nodes and 168 edges (undirected network) with a sparse spatial structural relationship.
\begin{table}[!h]
    \centering
    \caption{Datasets description.}
    \label{tab:dataset}
\resizebox{\columnwidth}{!}{
\begin{tabular}{@{}c|ccc@{}}
\toprule
data type                                                                       & dataset       & \#sensors & time span                              \\ \midrule
\multirow{2}{*}{\begin{tabular}[c]{@{}c@{}}highway\\ traffic flow\end{tabular}} & PEMS07        & 883       & 05/01/2017-08/31/2017                  \\
                                                                                & PEMS08        & 170       & 07/01/2016-08/31/2016                  \\ \midrule
\multirow{2}{*}{\begin{tabular}[c]{@{}c@{}}metro\\ crowd flow\end{tabular}}     & HZME(inflow)  & 80        & \multirow{2}{*}{01/01/2019-01/26/2019} \\
                                                                                & HZME(outflow) & 80        &                                        \\ \bottomrule
\end{tabular}
}
\end{table}

% 实验结果表格
\begin{table*}[!h]
    \centering
    \caption{Performance comparison on the highway traffic flow and metro crowd flow datasets.}
    \label{tab:result}
\resizebox{0.85\textwidth}{!}{
\begin{tabular}{c|ccc|ccc}
\hline
Datasets (Highway Traffic Flow) & \multicolumn{3}{c|}{PEMS07}                                           & \multicolumn{3}{c}{PEMS08}                                            \\ \hline
Metrics                         & MAE                   & RMSE                  & MAPE(\%)              & MAE                   & RMSE                  & MAPE(\%)              \\ \hline
VAR (2003)                      & 101.2                 & 155.14                & 39.69                 & 22.32                 & 33.83                 & 14.47                 \\
SVR (1997)                      & 32.97 ± 0.98          & 50.15 ± 0.15          & 15.43 ± 1.22          & 23.25 ± 0.01          & 36.15 ± 0.02          & 14.71 ± 0.16          \\
LSTM (1997)                     & 29.71 ± 0.09          & 45.32 ± 0.27          & 14.14 ± 1.00          & 22.19 ± 0.13          & 33.59 ± 0.05          & 18.74 ± 2.79          \\
DCRNN (2018)                    & 23.60 ± 0.05          & 36.51 ± 0.05          & 10.28 ± 0.02          & 18.22 ± 0.06          & 28.29 ± 0.09          & 11.56 ± 0.04          \\
STGCN (2018)                    & 27.41 ± 0.45          & 41.02 ± 0.58          & 12.23 ± 0.38          & 18.04 ± 0.19          & 27.94 ± 0.18          & 11.16 ± 0.10          \\
ASTGCN (2019)                   & 25.98 ± 0.78          & 39.65 ± 0.89          & 11.84 ± 0.69          & 18.86 ± 0.41          & 28.55 ± 0.49          & 12.50 ± 0.66          \\
GWN (2019)                      & 21.22 ± 0.24          & 34.12 ± 0.18          & 9.07 ± 0.20           & 15.07 ± 0.17          & 23.85 ± 0.18          & 9.51 ± 0.22           \\
GMAN (2020)                     & 21.56 ± 0.26          & 34.97 ± 0.44          & 9.51 ± 0.16           & 15.33 ± 0.03          & 26.10 ± 0.28          & 10.97 ± 0.37          \\
AGCRN (2020)                    & 22.56 ± 0.33          & 36.18 ± 0.46          & 9.67 ± 0.14           & 16.26 ± 0.43          & 25.62 ± 0.56          & 10.33 ± 0.34          \\
STSGCN (2020)                   & 23.99 ± 0.14          & 39.32 ± 0.31          & 10.10 ± 0.08          & 17.10 ± 0.04          & 26.83 ± 0.06          & 10.90 ± 0.05          \\
MTGNN (2020)                    & 20.57 ± 0.61          & 33.54 ± 0.73          & 9.12 ± 0.13           & 15.52 ± 0.06          & 25.59 ± 0.29          & 13.56 ± 1.11          \\
STFGNN (2021)                   & 22.07 ± 0.11          & 35.80 ± 0.18          & 9.21 ± 0.07           & 16.64 ± 0.09          & 26.22 ± 0.15          & 10.60 ± 0.06          \\
STGODE (2021)                   & 22.89 ± 0.15          & 37.49 ± 0.07          & 10.10 ± 0.06          & 16.79 ± 0.02          & 26.05 ± 0.11          & 10.58 ± 0.08          \\
DMSTGCN (2021)                  & 20.77 ± 0.57          & 33.67 ± 0.54          & 8.94 ± 0.42           & 16.02 ± 0.10          & 26.00 ± 0.21          & 10.28 ± 0.08          \\
ASTGNN (2021)                   & 20.62 ± 0.12          & 34.00 ± 0.21          & 8.86 ± 0.10           & 15.00 ± 0.35          & 24.70 ± 0.53          & 9.50 ± 0.11           \\
ASTGNN(p) (2021)                & 19.26 ± 0.17          & 32.75 ± 0.25          & 8.54 ± 0.19           & 12.72 ± 0.09          & 22.60 ± 0.13          & 8.78 ± 0.20           \\ \hline
CorrSTN                         & 19.62 ± 0.05          & 33.11 ± 0.23          & 8.22 ± 0.06           & 14.27 ± 0.17          & 23.67 ± 0.15          & 9.32 ± 0.06           \\
CorrSTN(p)                      & \textbf{18.10 ± 0.10} & \textbf{31.61 ± 0.12} & \textbf{7.58 ± 0.12}  & \textbf{12.56 ± 0.03} & \textbf{22.43 ± 0.03} & \textbf{8.53 ± 0.09}  \\ \hline \hline
Datasets (Metro Crowd Flow)     & \multicolumn{3}{c|}{HZME(inflow)}                                     & \multicolumn{3}{c}{HZME(outflow)}                                     \\ \hline
Metrics                         & MAE                   & RMSE                  & MAPE(\%)              & MAE                   & RMSE                  & MAPE(\%)              \\ \hline
VAR (2003)                      & 17.65                 & 28.1                  & 58.07                 & 22.35                 & 37.96                 & 96.68                 \\
SVR (1997)                      & 21.94 ± 0.02          & 40.73 ± 0.02          & 49.40 ± 0.07          & 25.59 ± 0.12          & 50.07 ± 0.17          & 91.71 ± 3.18          \\
LSTM (1997)                     & 22.53 ± 0.51          & 39.33 ± 0.35          & 60.12 ± 2.44          & 26.18 ± 0.32          & 48.91 ± 0.45          & 103.06 ± 8.52         \\
DCRNN (2018)                    & 12.25 ± 0.13          & 20.91 ± 0.33          & 25.53 ± 0.38          & 18.02 ± 0.16          & 31.45 ± 0.39          & 66.98 ± 1.65          \\
STGCN (2018)                    & 12.88 ± 0.28          & 22.86 ± 0.39          & 29.66 ± 1.50          & 19.12 ± 0.23          & 33.12 ± 0.36          & 73.66 ± 1.49          \\
ASTGCN (2019)                   & 13.10 ± 0.47          & 23.23 ± 0.81          & 33.29 ± 3.63          & 19.35 ± 0.51          & 33.20 ± 1.07          & 88.75 ± 4.00          \\
GWN (2019)                      & 11.20 ± 0.11          & 19.73 ± 0.46          & 23.75 ± 0.71          & 17.50 ± 0.12          & 30.65 ± 0.41          & 73.65 ± 2.72          \\
GMAN (2020)                     & 11.35 ± 0.20          & 20.60 ± 0.33          & 26.85 ± 0.72          & 18.03 ± 0.11          & 32.51 ± 0.37          & 74.57 ± 0.45          \\
AGCRN (2020)                    & 11.86 ± 0.71          & 24.39 ± 0.73          & 30.93 ± 1.82          & 19.34 ± 1.27          & 33.85 ± 1.16          & 88.85 ± 0.48          \\
STSGCN (2020)                   & 12.85 ± 0.10          & 23.20 ± 0.38          & 28.02 ± 0.19          & 18.74 ± 0.13          & 33.12 ± 0.43          & 76.85 ± 1.01          \\
MTGNN (2020)                    & 11.99 ± 0.39          & 20.57 ± 0.55          & 26.87 ± 0.64          & 18.79 ± 0.80          & 32.27 ± 0.60          & 87.63 ± 3.84          \\
STFGNN (2021)                   & 13.12 ± 0.23          & 23.02 ± 0.37          & 30.67 ± 0.53          & 18.90 ± 0.18          & 34.12 ± 0.43          & 77.32 ± 2.33          \\
STGODE (2021)                   & 11.36 ± 0.06          & 22.02 ± 0.14          & 40.50 ± 1.01          & 19.43 ± 0.38          & 33.67 ± 0.64          & 89.90 ± 2.57          \\
DMSTGCN (2021)                  & 12.64 ± 0.28          & 21.79 ± 0.53          & 28.21 ± 0.75          & 18.52 ± 0.37          & 32.26 ± 0.84          & 77.08 ± 0.76          \\
ASTGNN (2021)                   & 11.46 ± 0.08          & 20.84 ± 0.25          & 24.42 ± 0.30          & 17.94 ± 0.11          & 31.91 ± 0.32          & 72.46 ± 2.42          \\
ASTGNN(p) (2021)                & 10.94 ± 0.04          & 18.89 ± 0.11          & 23.33 ± 0.14          & 17.47 ± 0.03          & 30.78 ± 0.08          & 70.52 ± 0.27          \\ \hline
CorrSTN                         & 11.20 ± 0.06          & 19.71 ± 0.14          & 24.04 ± 0.38          & 17.26 ± 0.08          & 30.66 ± 0.15          & 65.33 ± 0.26          \\
CorrSTN(p)                      & \textbf{10.39 ± 0.10} & \textbf{17.23 ± 0.26} & \textbf{22.93 ± 0.35} & \textbf{15.24 ± 0.37} & \textbf{26.34 ± 0.70} & \textbf{51.22 ± 2.64} \\ \hline
\end{tabular}
}
\end{table*}

% 实验结果对比图
 \begin{figure*}[!h]
 \centering
 \includegraphics[width=\textwidth]{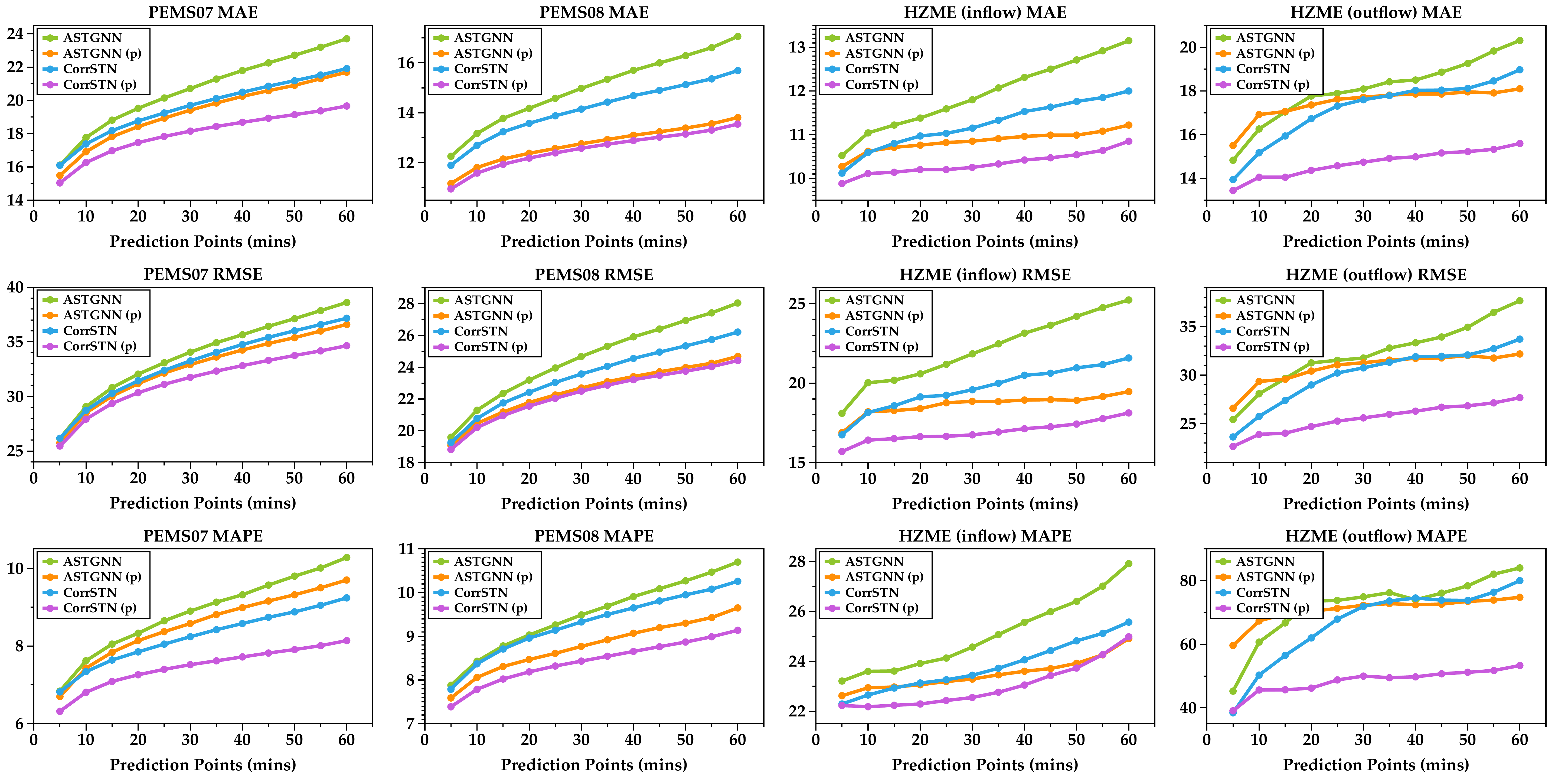}
 \caption{Performance comparsion of CorrSTN and ASTGNN as the prediction interval increases.}
 \label{fig:Experiments}
 \end{figure*}

\subsection{Performance}

We compare our model with fifteen baseline methods, including VAR~\cite{Lu:2003br}, SVR~\cite{Drucker:1997}, LSTM~\cite{Hochreiter:1997}, DCRNN~\cite{li2018diffusion}, STGCN~\cite{yu2018spatio}, ASTGCN~\cite{Guo.2019}, GWN~\cite{Wu:2019}, GMAN~\cite{Zheng.2020}, AGCRN~\cite{Bai:2020}, STSGCN~\cite{Song.2020}, MTGNN~\cite{Wu:2020}, STFGNN~\cite{Li_Zhu_2021}, STGODE~\cite{Fang:2021}, DMSTGCN~\cite{Han:2021} and ASTGNN~\cite{Guo.2021}. We repeat the experiments more than 5 times on all datasets to evaluate the performance, and the means and standard deviations are collected in~\tabref{tab:result}, where the bold font highlights the best values. The results obtained by using periodic data, including hourly, daily or weekly data, are marked by (p).

 With only hourly data, CorrSTN achieves superior performance in the metrics of MAE, RMSE, and MAPE on both the highway traffic flow and metro crowd flow datasets. We further enhance the predictive performance by using periodic data (hourly, daily or weekly) as input to our model. Thus, our model has the ability to outperform the state-of-the-art method ASTGNN(p). In the following, we will give detailed comparisons and analyses of these baselines and our model.

The traditional methods, VAR, SVR and LSTM, only consider the temporal features but ignore the spatial relationships among sensors. Thus, their capabilities of capturing spatiotemporal dependencies are quite limited. For example, the LSTM achieves the worst score in the metrics of MAE, RMSE and MAPE on the HZME (outflow) dataset, as shown in~\tabref{tab:result}.

The neural network-based methods (DCRNN, STGCN, ASTGCN, STSGCN, and ASTGNN) all take efforts to make use of spatiotemporal relationships in feature extraction. The DCRNN combines diffusion convolution and RNN to predict traffic flow. However, the ability for long-term forecasting is limited by the RNN capability. The STGCN, ASTGCN, STSGCN, and ASTGNN are four methods mixed of CNN-based and GNN-based components, where the temporal features are captured by the CNN, and the spatial features are captured by the GNN. Nevertheless, the critical correlation information is not taken into consideration in these methods. Due to the combination of correlation information into the GNN component, GWN, GMAN, AGCRN, MTGNN, STFGNN, STGODE, and DMSTGCN can obtain a more satisfactory performance for each dataset, as shown in~\tabref{tab:result}. The main reasons are that the adaptive graph structure is built with training parameters (e.g., GWN, GMAN, and AGCRN), a sparse graph matrix is constructed with the DTW algorithm (e.g., STFGNN and STGODE), or a dynamic graph convolution is employed to capture neighbor sensor features (e.g., MTGNN and DMSTGCN).

However, the graph structures of the above methods cannot adequately represent similar patterns among various sensors. To address the problem, we use the MIC algorithm to detect more widespread relationships for combining similar features in the CIGNN component. As a result, CorrSTN outperforms AGCRN on the HZME (flow) dataset by 21.2\%, 22.2\% and 42.4\% in the metrics of MAE, RMSE and MAPE, respectively, as shown in~\tabref{tab:result} (AGCRN (2020)). On the other hand, unlike dynamic graph convolution (e.g., MTGNN and DMSTGCN), we employ a spatial dynamic weight matrix to fit the varying changes throughout the training and test processes. Compared with DMSTGCN on the HZME (flow) dataset, our model improves the performance by 17.7\%, 18.4\% and 33.5\% in the metrics of MAE, RMSE and MAPE, respectively, as shown in~\tabref{tab:result} (DMSTGCN (2021)).

For the attention mechanism, several previous methods suffer from unwanted distractions due to vanilla attention, such as spatiotemporal attention (GMAN) and trend-aware attention (ASTGNN). In our model, with the help of SCorr, CIATT can aggregate similar patterns from different sensors and apply more focused attention weights to match the most relevant sequence pattern. Thus, compared with GMAN on the HZME (flow) dataset, the performance of CorrSTN can be enhanced by 15.5\%, 19.0\% and 31.3\% in the metrics of MAE, RMSE and MAPE, respectively, as shown in~\tabref{tab:result} (GMAN (2020)).

The data selection scheme for periodic data is a crucial part to enhance model performance. Instead of the traditional exhaustive search, we design an appropriate data selection scheme by using the efficient representation TCorr. In contrast to ASTGNN (p), which adopts the hourly and weekly data, our CorrSTN (p) adopts the hourly and daily data as input for model training on the HZME (outflow) dataset. In this manner, we enhance the model CorrSTN (p) and make improvements by 12.7\%, 14.4\% and 27.4\% in the metrics of MAE, RMSE and MAPE, respectively, as shown in~\tabref{tab:result} (ASTGNN (p) (2021)).

To show the stability of our model, we present an illustration at each predicted point of the CorrSTN and ASTGNN to clearly show our improvement, as shown in~\figref{fig:Experiments}. On the PEMS07, HZME (inflow) and HZME (outflow) datasets, our model makes notable advancements compared with ASTGNN. Especially for long-term forecasting, our model achieves a more stable predictive performance, as shown in~\figref{fig:Experiments} (e.g., PEMS07 (MAE, MAPE), HZME (inflow) (MAE, RMSE) and HZME (outflow) (MAE, RMSE, MAPE)).

Overall, our work offers more accurate results for traffic flow forecasting. The improvement is attributed to the accurate aggregation in CIGNN and the focused attention weights in CIATT with SCorr. Moreover, based on TCorr, a novel method searching for an effective scheme further improves our model performance.

%an effective scheme is {\color{red}proposed} to further improve our model's performance.}}

\subsection{Ablation Experiments}

\begin{figure}[!h]
\centering
\includegraphics[width=\columnwidth]{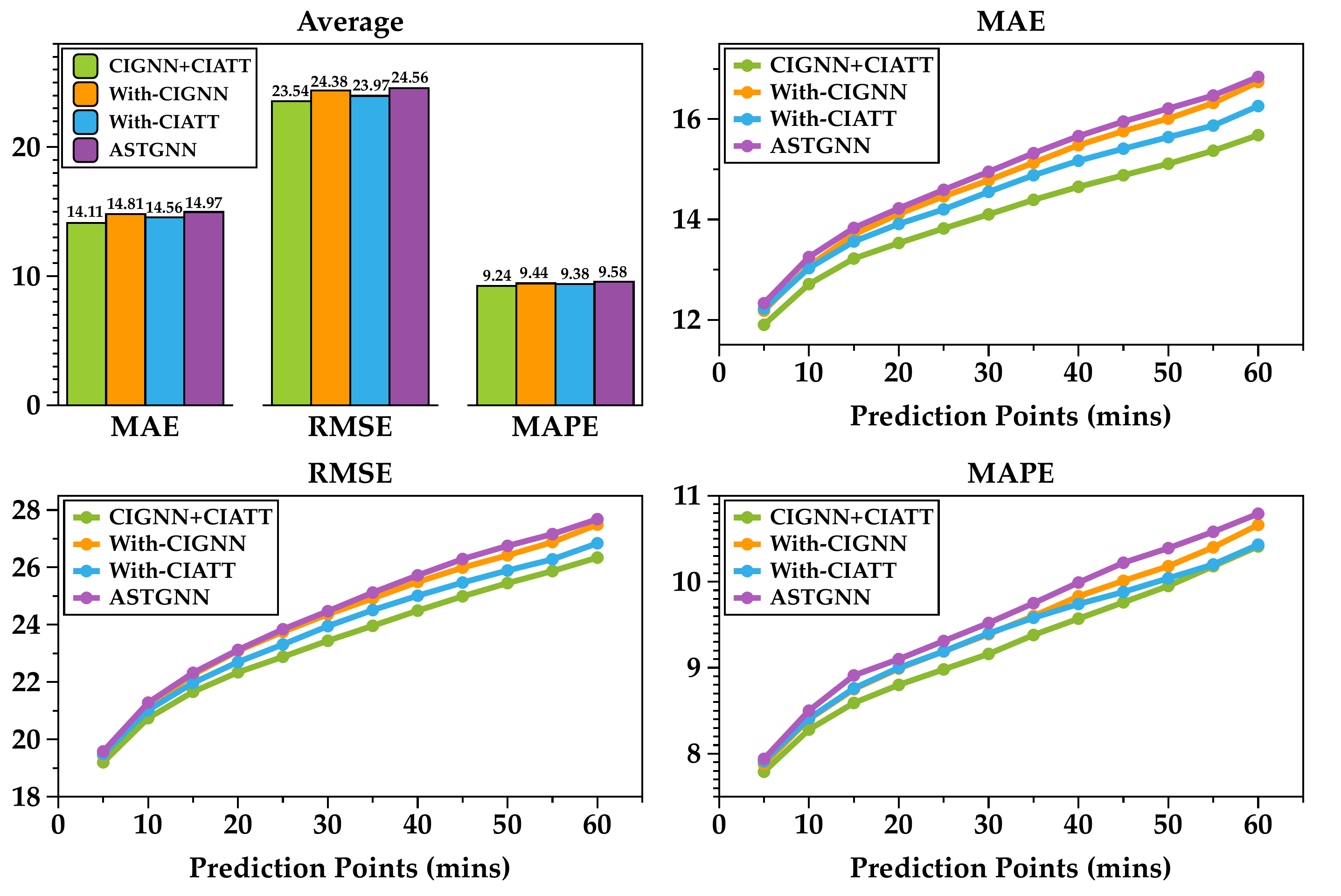}
\caption{Ablation results on the PEMS08 dataset.}
\label{fig:pems08}
\end{figure}

In this subsection, we verify the effectiveness of our CIGNN and CIATT components on the PEMS08 dataset. For comparison, we assign the state-of-the-art ASTGNN (w/o CIGNN \& CIATT) as the baseline. And, $with$-CIGNN ($with$-CIATT) denotes the model using only the CIGNN (CIATT) component, and CIGNN+CIATT denotes the model using the CIGNN and CIATT components. Our model takes the hourly data as the input for training and shows the prediction results for all prediction points, as shown in \figref{fig:pems08}. 

From the experimental results, we find that $with$-CIGNN (resp. $with$-CIATT) outperforms the baseline by 1.04\%, 0.72\% and 1.50\% (resp. 2.74\%, 2.42\% and 2.13\%) in the metrics of MAE, RMSE and MAPE, respectively. Moreover, CIGNN+CIATT outperforms the baseline by 5.72\%, 4.16\% and 3.61\% in the metrics of MAE, RMSE and MAPE, respectively. Thus, the CIGNN and CIATT modules can make significant increases in prediction accuracy. We also find that the combined contribution (CIGNN+CIATT) is larger than the sum of the contributions of CIGNN and CIATT ($with$-CIGNN and $with$-CIATT). This shows that our CIGNN and CIATT are able to interact and positively influence each other by using the correlation information between spatial and temporal features.

%by 

%based on the correlation information between spatial and temporal features, .}

%Therefore, by combining CIGNN and CIATT together, we can achieve superior predictive performance.

Furthermore, comparing the results, we find that the CIATT component shows higher effectiveness than the CIGNN component in the traffic flow forecasting task. In our model, the CIATT component focuses the attention weights on the most relevant sequence and constructs more correct features for the other components, as shown in \figref{fig:CIATT-diff}. Thus, a powerful improvement can be achieved by our CIATT component.

\subsection{Time Cost Study}
We implement the proposed model in Python 3.8 and PyTorch 1.7.0. The model has been successfully executed and tested on the Linux platform with an Intel (R) Xeon (R) Gold 6240R CPU@2.40 GHz and NVIDIA TESLA V100 (PCI-E) GPU 32 GB card. We list the time cost of SCorr and TCorr (CPU: AMD 3970x 32C64T). Although the time cost of Scorr is expensive, the cost can be reduced with more parallel computing cores. Furthermore, the cost is only needed once for each dataset. In addition, as shown in~\tabref{tab:time}, we can search for an appropriate data scheme in 18s on the PEMS07 dataset. However, the neural networks cannot finish their one epoch training for the traditional exhaustive search at the same time.

\begin{table}[!h]
    \centering
    \caption{Training and test time cost on four datasets.}
    \label{tab:time}
\resizebox{0.8\columnwidth}{!}{
\begin{tabular}{@{}ccc@{}}
\toprule
              & SCorr Time Cost & TCorr Time Cost     \\ \midrule
PEMS07        & 04:42:00        & 00:00:18            \\
PEMS08        & 00:20:00        & 00:00:02            \\
HZME(inflow)  & 00:00:17        & \textless{}00:00:01 \\
HZME(outflow) & 00:00:17        & \textless{}00:00:01 \\ \bottomrule
\end{tabular}
}
\end{table}

\section{Comparison and Analysis Beyond Performance}
In this section, we perform experiments to discuss and analyze the data selection schemes, dynamic SCorr and the influence of different top-$U$.

%efficiency of TCorr. Moreover, we give detailed discussions about hyperparameter top-$U$ and SCorr.

\subsection{Data Selection Schemes}
\label{sec:TCorr}
%In this subsection, we present temporal correlation information representations on the four real-world datasets. Then, for the different representations, we explain the reasons based on the background of the datasets.

\begin{figure}[!h]
\centering
\includegraphics[width=\columnwidth]{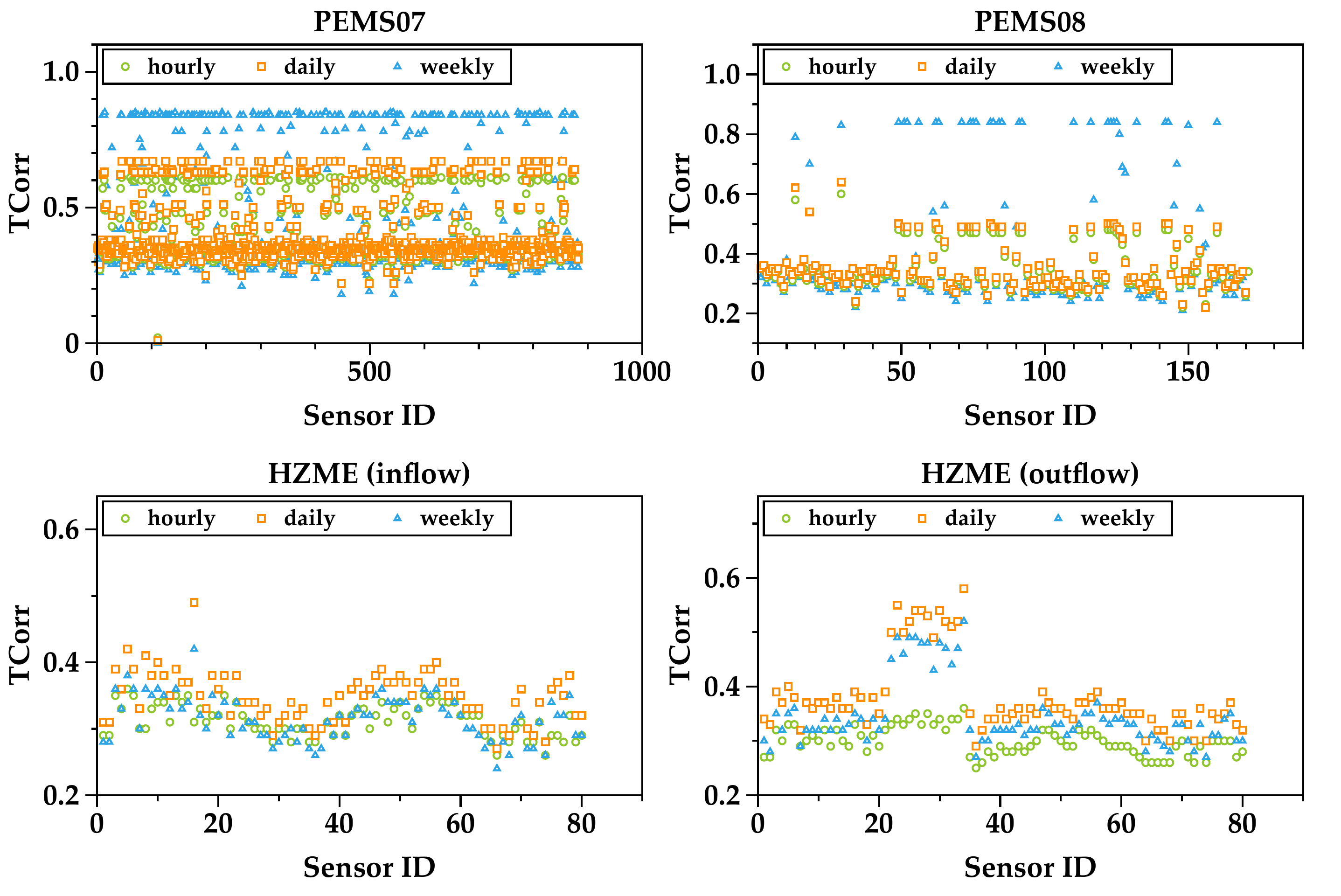}
\caption{Temporal correlation information of each sensor on four datasets.}
\label{fig:TCorr-datanode}
\end{figure}

In this subsection, we consider the four real-world datasets. Different datasets have different temporal correlation information representations, as shown in~\figref{fig:TCorr-datanode}. For the sake of analysis, we sum and average the scores of all sensors, as shown in~\tabref{tab:TCorr-avg}. It is obvious that the weekly data have the highest correlation degree with the prediction data on the highway traffic flow datasets (PEMS07 and PEMS08). In contrast, the daily data have the highest correlation degree with the prediction data on the metro traffic flow datasets (HZME (inflow) and HZME (outflow)).

% TCorr 节点分布

\begin{table}[h]
    \centering
    \caption{Temporal correlation information (average) on four datasets.}
    \label{tab:TCorr-avg}
\resizebox{\columnwidth}{!}{
\begin{tabular}{@{}ccccc@{}}
\toprule
\multicolumn{1}{l}{}        & PEMS07         & PEMS08         & HZME(inflow)   & HZME(outflow)  \\ \midrule
\multicolumn{1}{c|}{hourly} & 0.379          & 0.343          & 0.312          & 0.300          \\
\multicolumn{1}{c|}{daily}  & 0.397*         & 0.356*         & \textbf{0.348} & \textbf{0.380} \\
\multicolumn{1}{c|}{weekly} & \textbf{0.426} & \textbf{0.411} & 0.316*         & 0.346*         \\ \bottomrule
\end{tabular}
}
\end{table}

%Thus, the different characteristics of traffic flow are reflected in TCorr.

It is reasonable that these four datasets have different temporal correlation information representations. As stated in~\secref{sec:datasets}, the highway traffic flow datasets (PEMS07 and PEMS08) and the metro crowd flow datasets (HZME (inflow) and HZME (outflow)) are collected from different locations and environments. Concretely, highway traffic flow datasets are collected from highway roads in California, which reflect the characteristics of long-distance travel. In contrast, the metro crowd flow datasets are collected from the metro in Hangzhou, which reflects the characteristics of short-distance travel.

% \subsection{Data Selection Schemes}
% \label{sec:TCorr-scheme}

Hence, for the different representations, we need to design different schemes, as stated in~\secref{sec:TCorr-define}. In the following, we will verify the model performance with different data selection schemes on the HZME (outflow) dataset. We conduct experiments to compare the performance of our scheme and traditional exhaustive search schemes. The type of scheme is designed as scheme $h_{i}d_{j}w_{k}$, as shown in~\tabref{tab:schemes}. The $h_1$ ($h_0$), $d_1$ ($d_0$), and $w_1$ ($w_0$) denote that the input data (do not) contain hourly, daily and weekly data, respectively.

\begin{table}[h]
    \centering
    \caption{The seven data selection schemes on the HZME (outflow) datasets.}
    \label{tab:schemes}
\resizebox{0.75\columnwidth}{!}{
\begin{tabular}{@{}cccc@{}}
\toprule
scheme                      & hourly data  & daily data  & weekly data  \\ \midrule
\multicolumn{1}{c|}{$h_1d_0w_0$} & $\surd$ &         &         \\
\multicolumn{1}{c|}{$h_0d_1w_0$} &         & $\surd$ &         \\
\multicolumn{1}{c|}{$h_0d_0w_1$} &         &         & $\surd$ \\
\multicolumn{1}{c|}{$h_1d_1w_0$} & $\surd$ & $\surd$ &         \\
\multicolumn{1}{c|}{$h_1d_0w_1$} & $\surd$ &         & $\surd$ \\
\multicolumn{1}{c|}{$h_0d_1w_1$} &              & $\surd$ & $\surd$ \\
\multicolumn{1}{c|}{$h_1d_1w_1$} & $\surd$ & $\surd$ & $\surd$ \\ \bottomrule
\end{tabular}
}
\end{table}

%{\color{blue}To verify the performance of each scheme, we conduct experiments with different input data. Then, we illustrate the average performance and each prediction point performance in the metrics of MAE, RMSE and MAPE, as shown in~\figref{fig:TCorr-outflow}. With our $h1d1w0$ scheme, our model achieves outstanding performance at all prediction points.}

%With the help of TCorr, we adopt the $h1d1w0$ scheme as the appropriate data selection scheme to enhance predictive performance.

\begin{figure}[!h]
\centering
\includegraphics[width=\columnwidth]{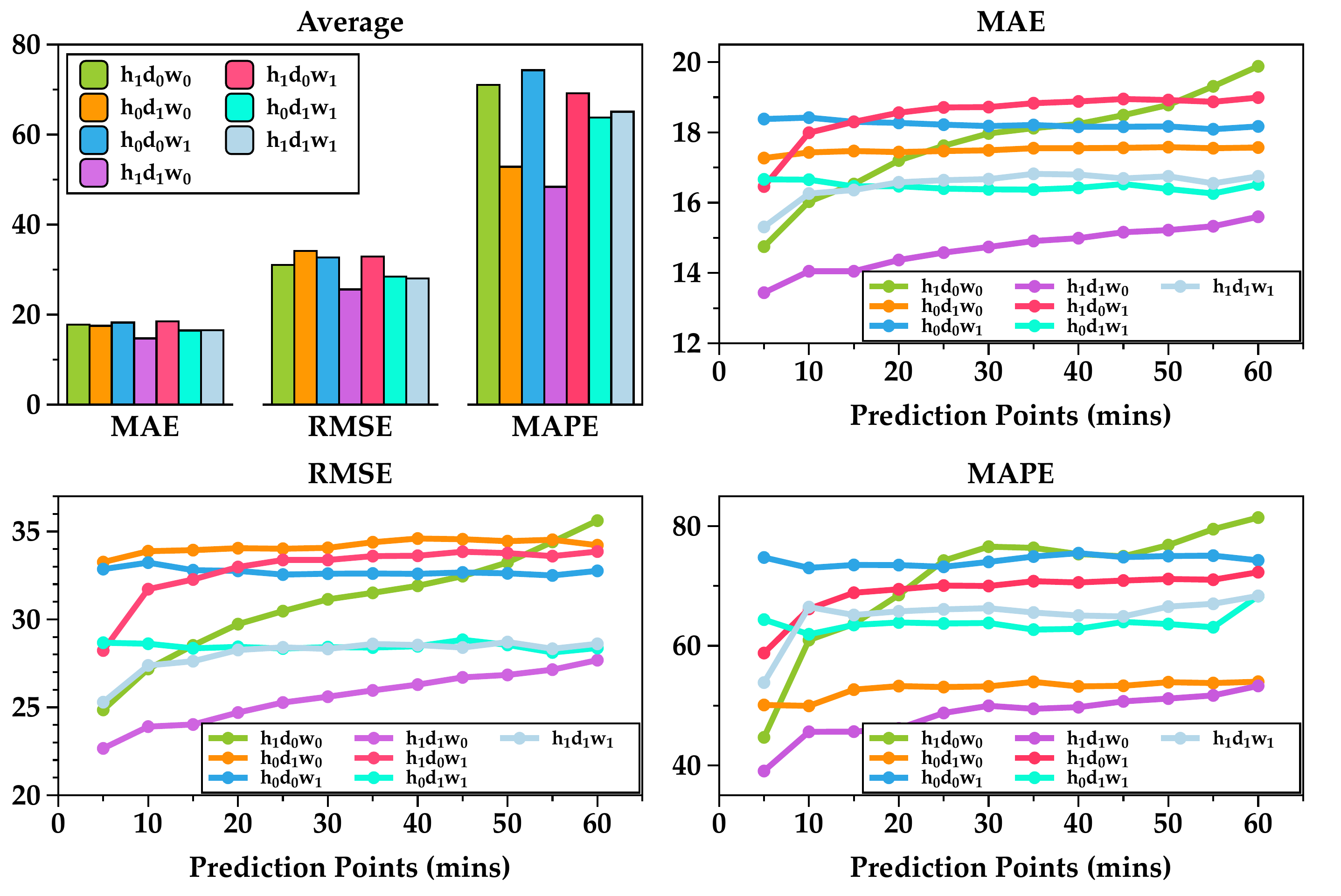}
\caption{Results on the HZME (outflow) dataset with six different data schemes.}
\label{fig:TCorr-outflow}
\end{figure}

% with some comparisons

The average performance and each prediction point performance in the metrics of MAE, RMSE and MAPE are shown in~\figref{fig:TCorr-outflow}. First, we analyze the effect of hourly data. Compared with the $h_0d_1w_0$ (or $h_0d_0w_1$) scheme, the $h_1d_0w_0$ scheme gives rise to the predictive performance at the short-term prediction points. However, the performance sharply decreases at the long-term prediction points with the $h_1d_0w_0$ scheme. It shows that hourly data only reflect the short-term trend in the HZME (outflow) dataset. Moreover, by comparing the $h_0d_1w_1$ and $h_1d_1w_1$ schemes, we can also observe the improvement by hourly data at the short-term prediction points. Then, we consider the daily and weekly data by comparing the $h_1d_1w_0$, $h_1d_0w_1$ and $h_1d_1w_1$ schemes. The experimental results demonstrate that the daily data can improve the model performance at short-term and long-term prediction points. In contrast, the weekly data decrease the model performance. Thus, we can conclude that the daily data make more improvements in predictive performance than the weekly data. Moreover, as shown in~\tabref{tab:TCorr-avg}, the result of $\Delta_{dw}<0$ demonstrates that there is no regular weekly periodic property in the HZME (outflow) dataset.

%In the case study of~\secref{sec:TCorr-case-study}, we also illustrate the traffic flow sequence to show the periodic property, as shown in~\figref{fig:TCorr-show}.

%The computation costs are quite expensive to select the best scheme based on all results. 

Overall, with the $h_1d_1w_0$ scheme, our model can achieve outstanding performance at all prediction points. Compared with the traditional exhaustive search method, our method with TCorr can effectively search for an appropriate data selection scheme without trying all possible schemes. With the appropriate scheme, we can further improve the model performance for traffic flow forecasting tasks.

%Thus, we can avoid the expensive computation costs in this way

%{\color{black}Although, based on all results, of the seven schemes, we can determine that the $h1d1w0$ scheme is the best scheme for the HZME (outflow) dataset. However, it should be noted that the computation costs are quite expensive.} 

%{\color{blue}Overall, we make an effective method to search for the data selection scheme. With the scheme, {\color{black}we are able to further improve the model performance for traffic flow forecasting tasks.}}

\subsection{Static and Dynamic SCorr}
\label{sec:sta-dyn}
In this subsection, we compare the effect of static and dynamic SCorr on the HZME (inflow) dataset. The dynamic SCorr is calculated by cyclic forwarding along the timeline as~\equref{equ:SCorr} with $T$ at 12 timestamps (one hour). 

\begin{figure}[h]
\centering
\includegraphics[width=\columnwidth]{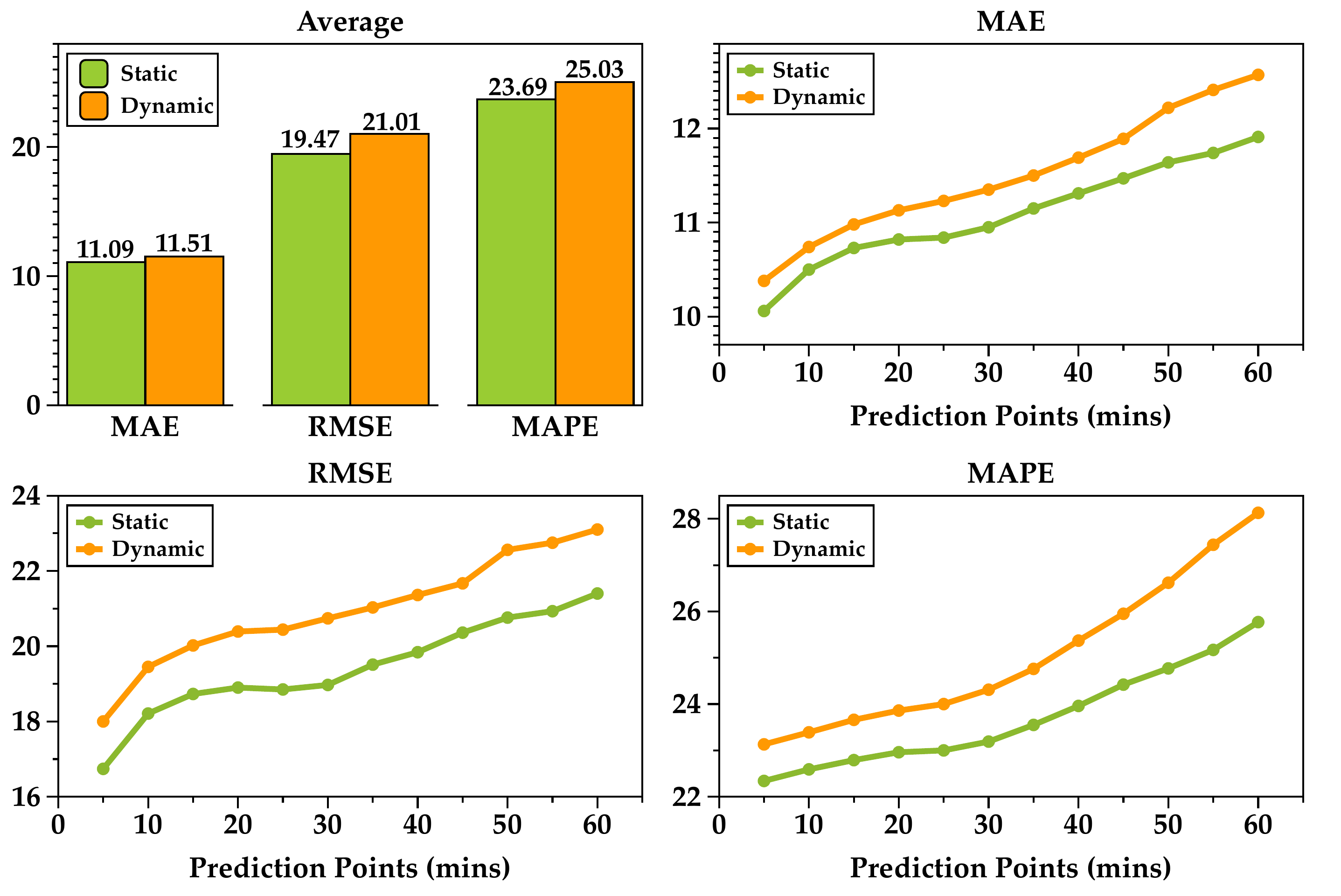}
\caption{Results on the HZME (inflow) dataset with static and dynamic spatial correlation information.}
\label{fig:static-dynamic}
\end{figure}

As shown in~\figref{fig:static-dynamic}, compared with CorrSTN with dynamic SCorr, CorrSTN with static SCorr achieves markedly better performance in the metrics of MAE, RMSE and MAPE. Although the dynamic SCorr contains rich information at each time point, the high-frequency change makes it difficult for the model to fit all training data. In fact, the correlative relationships do not frequently change. Thus, the dynamic SCorr leads to a performance decrease. Moreover, we use the spatial dynamic weight matrix to adaptively adjust the relationships in the CIGNN component. Therefore, we can also dynamically control the feature construction during the prediction process.

\subsection{CIATT with Different Top U}
In this subsection, we adopt six different $U$ hyperparameters to compare the performance with different levels of spatial correlation information on the HEME (inflow) dataset. We set the values of $U$ at 2, 3, 4, 5 and 8, and the other hyperparameters are set as shown in~\tabref{tab:setting}.

\begin{figure}[h]
\centering
\includegraphics[width=\columnwidth]{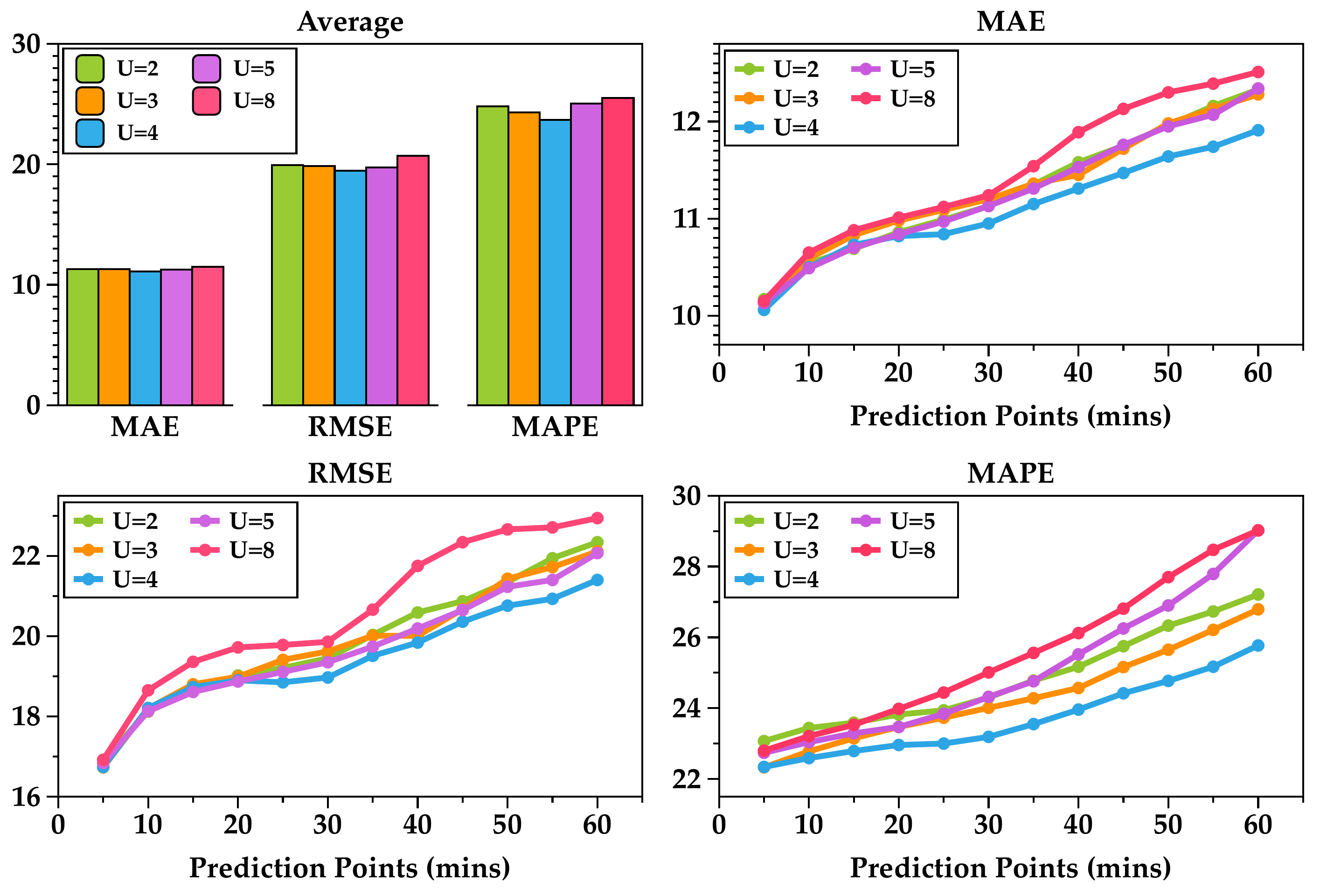}
\caption{Results on the HZME (inflow) dataset with six different $U$ schemes.}
\label{fig:different-topk}
\end{figure}

As shown in~\figref{fig:different-topk}, we find that the CorrSTN model achieves the best performance when $U$ is set at 4 on the HEME (inflow) dataset. Moreover, the performance can be improved as $U$ increases from $U=2$ until $U=4$, and the peak performance is achieved at $U=4$. We also find that the CIATT component is affected by the hyperparameter $U$. Concretely, as $U$ increases, the increasing number of relevant sequences can improve the predictive performance at the beginning. However, after achieving the best performance, the performance decreases due to the increasing number of irrelevant features (see $U=5$ and $U=8$ in~\figref{fig:different-topk}). Therefore, with an appropriate hyperparameter $U$, CIATT can further improve the predictive performance.

%Thus, the model can achieve improvements in predictive performance. However, 

%we set different hyperparameters $U$ for different experiments, as shown in~\tabref{tab:setting}.

%According to the experimental results, we deduce that the dynamic SCorr is not suitable in CIATT. The spatial dynamic weight matrix can 

\section{Case Study}
In this section, we conduct case studies to show that SCorr and TCorr can accurately represent the data correlations. Moreover, we also provide visual comparisons to clearly demonstrate our improvement in predictive performance.

%we provide details of SCorr and TCorr on four real-world datasets. Moreover, we present the predicted sequences of CorrSTN and ASTGNN for visual comparisons.

%We conduct several case studies to show that our approach
%can improve the fairness of news recommendation results.
%We randomly select a male user and a female user, and predict
%the ranking scores of candidate news based on their
%clicked news using NRMS and FairRec.
%
%We also conduct a case study to show the learned communities.
%We do statistical analyses on the communities, two
%of which are presented in Figure 6.

\begin{figure*}[t]
\centering
\includegraphics[width=0.95\textwidth]{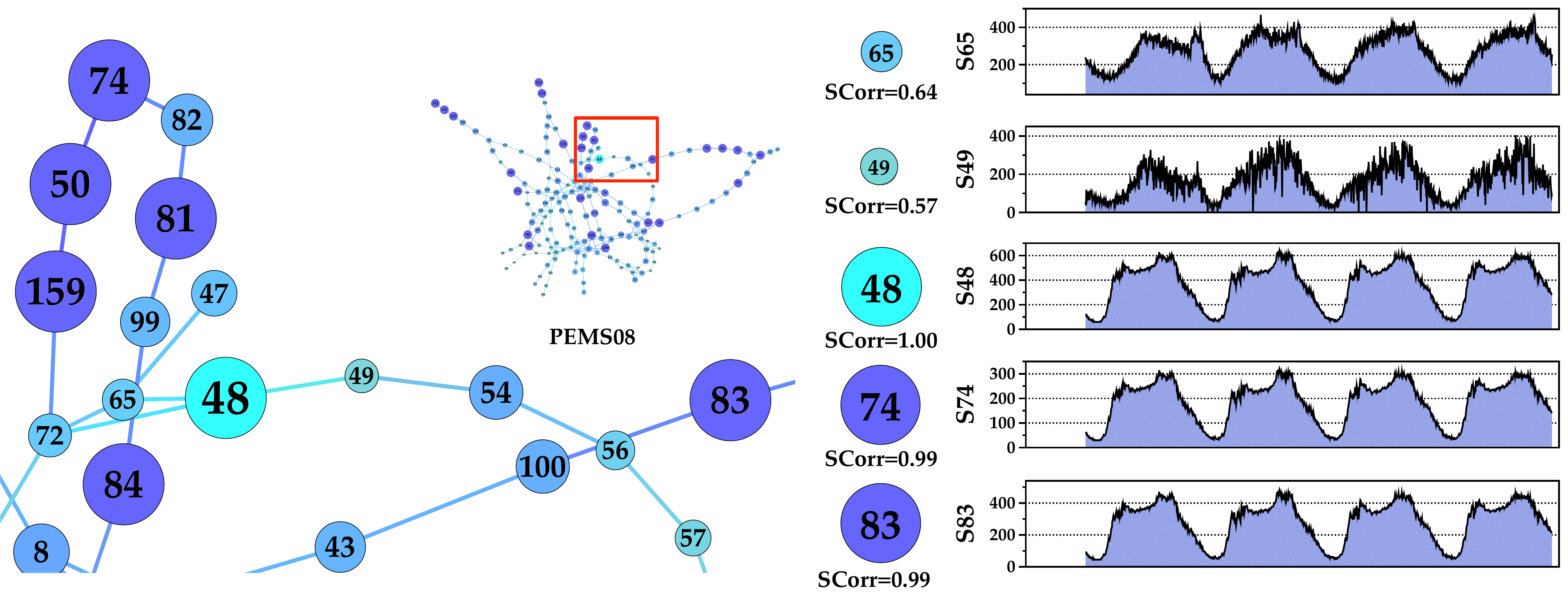}
\caption{The spatial correlation information between sensor $48$ and other sensors on the PEMS08 dataset.}
\label{fig:DataCorrelation}
\end{figure*}

\subsection{The Effect of SCorr}
In the traffic flow data, correlation information exists among sensors in different traffic datasets. Taking the PEMS08 dataset as an example, we calculate the correlation of sensors by using SCorr, as shown in \figref{fig:DataCorrelation}. The size of each sensor denotes its correlation degree with sensor 48. Note that both sensor 74 and sensor 83 have strong correlation representations with sensor 48, although there are no structural paths among them directly. The counterintuitive phenomenon shows that correlation information is more significant than structural information in discovering spatiotemporal dependencies and dynamic relationships in traffic forecasting tasks. Furthermore, according to the sensor 48 sequence and four other sequences, as shown in \figref{fig:DataCorrelation}, we find that SCorr can capture widespread associations accurately and provide reliable representations for evaluating the correlation strength.

\begin{figure}[t]
\centering
\includegraphics[width=\columnwidth]{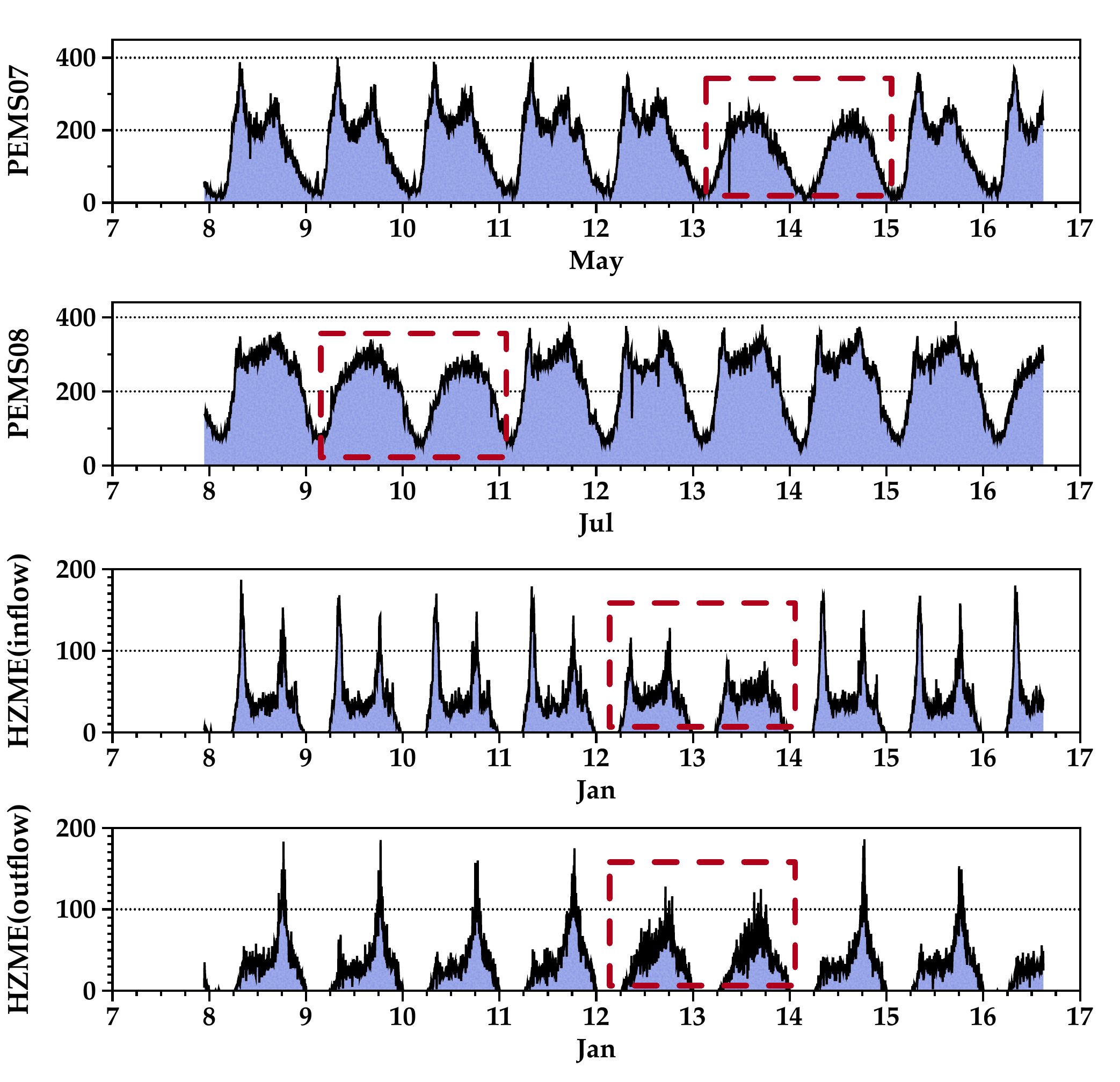}
\caption{The original data on the PEMS07, PEMS08, HZME (inflow) and HZME (outflow) datasets.}
\label{fig:TCorr-show}
\end{figure}

 \begin{figure*}[!t]
 \centering
 \includegraphics[width=0.85\textwidth]{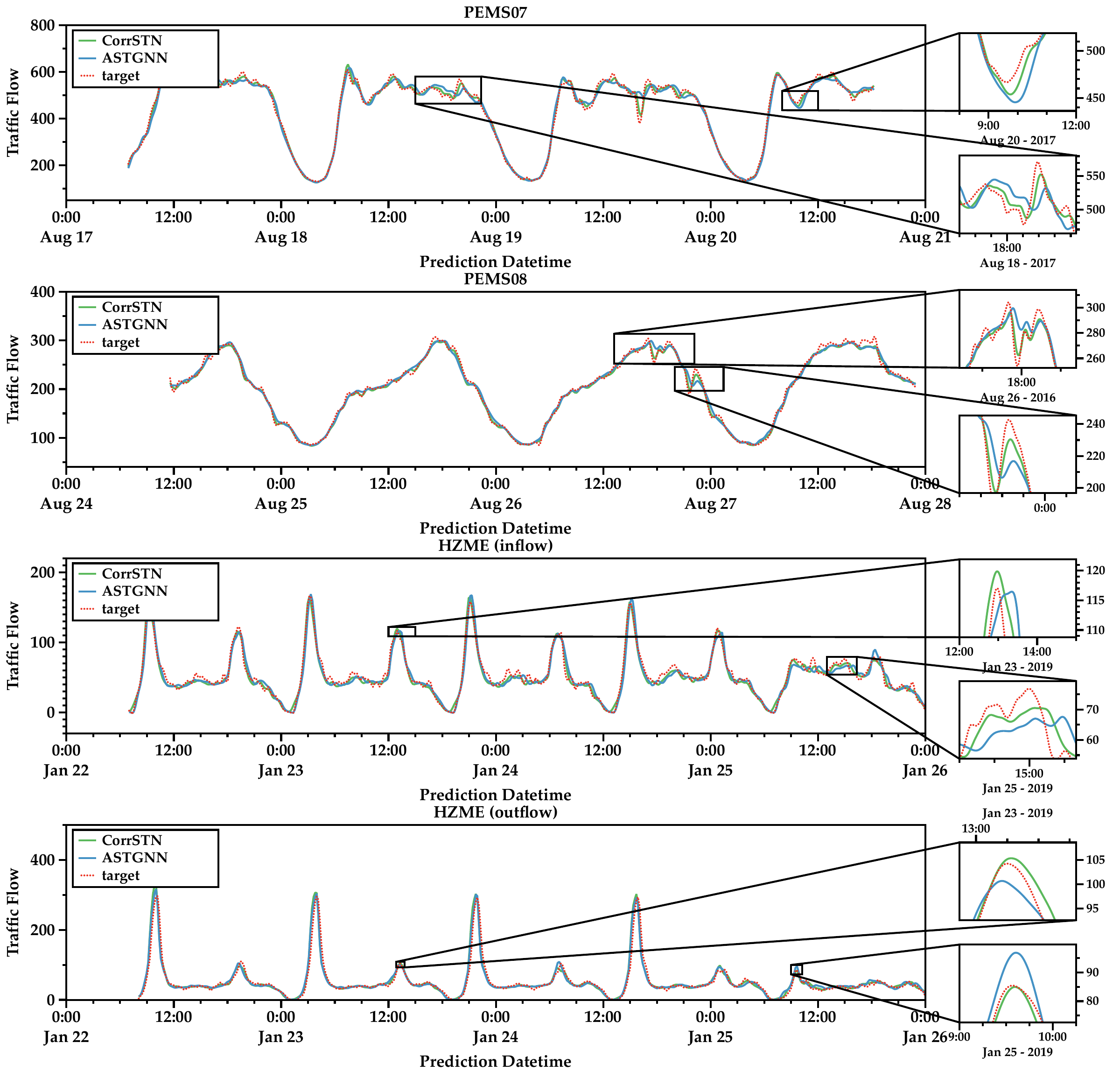}
 \caption{Visualizations on four real-world datasets.}
 \label{fig:case}
 \end{figure*}

\subsection{The Effect of TCorr}
\label{sec:TCorr-case-study}
%We illustrate the original data to show the dataset characteristics, as shown in \figref{fig:TCorr-show}. 

We take a part of the data as an example to show the effect of TCorr on four real-world datasets, as shown in \figref{fig:TCorr-show}. We can see that the PEMS07 and PEMS08 datasets have a stronger correlation with the weekly data, as shown on May 13-14 (weekend) in the PEMS07 dataset and Jul 9-10 (weekend) in the PEMS08 dataset. Compared with other days, the two weekends are special days in each week, which do not have morning peaks or evening peaks. It is demonstrated that the regular weekly periodic property exists in the two datasets. Meanwhile, the other days also show the regular daily periodic property. In contrast, the HZME inflow and outflow datasets strongly correlate with daily data, as shown on Jan 12-13 (weekend) on the two datasets. Compared with workdays, weekends also show similar sequences in the morning peaks and evening peaks. It is shown that all days have a regular daily periodic property. 

Based on the case study, we find that the effect of TCorr accurately corresponds with the data characteristics. Additionally, TCorr is effective for detecting the periodic property.

%According to a result, the scores of TCorr show the PEMS07 and PEMS08 datasets have a stronger weekly correlation. On the contrary, the HZME inflow and outflow datasets have a stronger daily correlation. 

% than the daily correlation

%Based on the comparisons, we prove that the TCorr representation is effective for detecting the periodic property.

%Thus, the different characteristics prove the correctness of TCorr and the effectiveness of different data selection schemes.

\subsection{Visual Comparisons}
We visualize the forecasting results of the CorrSTN and ASTGNN models, as shown in \figref{fig:case}. To clearly show the improvement, we smooth all sequences by the locally estimated scatterplot smoothing (LOESS) method~\cite{Cleveland:1979}, which is a nonparametric regression method.

The visualization results show that our model can fit the target sequences better than the ASTGNN model. Even though the target sequences are hard to fit, we can also obtain satisfactory predicted results, for example, the magnified sequences on Aug 18, 2017 of the PEMS07 dataset and Aug 26, 2016 of the PEMS08 dataset. Moreover, CorrSTN can achieve better performance at the sequences of peaks and troughs, for example, the magnified sequences on Aug 20, 2017 of the PEMS07 dataset and Jan 25, 2019 of the HZME (outflow) dataset.

%Utilizing TCorr, we explore the correlation pattern among different periodic data to identify the most relevant data, and then design an efficient data selection scheme to further enhance model performance.

%%%%%%%%%%%%%%% Conclusion %%%%%%%%%%%%%%%
\section{Conclusion}
In this paper, we propose an effective neural network-based network CorrSTN to predict traffic flow data in intelligent transportation systems. Considering the correlation information deeply, we first propose two elaborate spatiotemporal representations named SCorr and TCorr for spatiotemporal sensor sequences. Then, by using SCorr, we design two crucial components named CIGNN and CIATT in our CorrSTN model. The CIGNN component can improve the feature aggregation efficiency and produce more correct features. The CIATT component can construct more focused attention weights to extract features from relevant sequences. We use TCorr to mine the correlations among different periodic data to and design an effective data scheme for periodic datasets. Finally, we conduct experiments to compare CorrSTN with fifteen baseline methods on the highway traffic flow and metro crowd flow datasets. The experimental results demonstrate that our CorrSTN outperforms the state-of-the-art methods in terms of predictive performance. In particular, on the HZME (outflow) dataset, our model makes significant improvements compared with the ASTGNN model by 12.7\%, 14.4\% and 27.4\% in the metrics of MAE, RMSE and MAPE, respectively. Moreover, the case studies show that SCorr can elaborately present the spatiotemporal features, while TCorr can help to select the appropriate data schemes.

\backmatter

\bmhead{Acknowledgments}
%\section*{Acknowledgement}
This research is supported by the National Key R\&D Program of China (No. 2021ZD0113002), National Natural Science Foundation of China (No. 61572005, 62072292, 61771058) and Fundamental Research Funds for the Central Universities of China (No. 2020YJS032). The support and resources from the Center for High Performance Computing at Beijing Jiaotong University are also gratefully acknowledged.

\section*{Declarations}
\textbf{Conflict of interest} The authors declare that they have no known competing ﬁnancial interests or personal relationships that could have appeared to inﬂuence the work reported in this paper.

\noindent\textbf{Data Availability} The datasets, code and pre-trained models generated and analysed during the current study are available in the CorrSTN repository, https://github.com/bjtu-ccd-lab/CorrSTN.

%%===========================================================================================%%
%% If you are submitting to one of the Nature Portfolio journals, using the eJP submission   %%
%% system, please include the references within the manuscript file itself. You may do this  %%
%% by copying the reference list from your .bbl file, paste it into the main manuscript .tex %%
%% file, and delete the associated \verb+\bibliography+ commands.                            %%
%%===========================================================================================%%

\bibliography{CorrSTN.bib}
%% if required, the content of .bbl file can be included here once bbl is generated
%%\input sn-article.bbl

%% Default %%
%%\input sn-sample-bib.tex%

\end{document}